\documentclass[runningheads]{llncs}

\usepackage{eccv}

\usepackage{eccvabbrv}

\usepackage{graphicx}
\usepackage{booktabs}
\usepackage{algorithm}
\usepackage{algorithmic}
\usepackage{amsmath}
\usepackage{graphicx}
\usepackage{multirow}
\usepackage{array}
\usepackage{vcell}
\usepackage{geometry}
\usepackage{pgfplots}
\usepackage[table]{xcolor}
\usepackage[accsupp]{axessibility}  
\usepackage{wrapfig}
\usepackage{ragged2e}
\usepgfplotslibrary{groupplots} 

\usepackage{hyperref}

\usepackage{orcidlink}
\usepackage{tikz}

\usepackage{ulem}
\usepgfplotslibrary{groupplots}
\usetikzlibrary{spy}

\pgfplotsset{compat=1.18}
\newcolumntype{H}{>{\setbox0=\hbox\bgroup}c<{\egroup}@{}}

\definecolor{CBBlue}{RGB}{0,114,178}
\definecolor{CBOrange}{RGB}{230,159,0}
\definecolor{CBGreen}{RGB}{56, 128, 15}
\definecolor{CBRed}{RGB}{213, 94, 0}
\definecolor{CBPink}{RGB}{204, 121, 167}
\definecolor{CBCyan}{RGB}{86, 180, 180}

\definecolor{MPLRed}{HTML}{d62728}
\definecolor{MPLGreen}{HTML}{2ca02c}
\definecolor{MPLOrange}{HTML}{ff7f0e}
\definecolor{MPLBlue}{HTML}{1f77b4}

\newcommand{\nn}[1]{{#1}}
\newcommand{\ls}[1]{{#1}}

\begin{document}

\title{Defending from GeoLocalization through Adversarial Road Trips}

\author{Niccolò Niccoli\inst{1}\orcidlink{0009-0000-9311-7862} \and
Federico Becattini\inst{2}\orcidlink{0000-0003-2537-2700} \and
Lorenzo Seidenari\inst{1}\orcidlink{0000-0003-4816-0268}}

\authorrunning{N.~Niccoli et al.}


\institute{
University of Florence, Florence, Italy\\
\email{name.surname@unifi.it}
\and
University of Siena, Siena, Italy\\
\email{name.surname@unisi.it}
}

\maketitle

\begin{abstract}

Retrieval-based image geolocalization has emerged as a powerful technique for determining the location of a query image by matching it against a large, geotagged database. The success of deep learning based approaches has raised concerns regarding privacy and safety. A way to protect users from geolocalization is to design adversarial attacks for such methods. In this paper, we introduce RoadTrip Attack (RTA), a novel and highly effective targeted adversarial attack for geolocalization.  RTA conceptualizes the adversarial process as finding an optimal ``distractor'' journey to a specific, attacker-chosen location. It employs a beam search algorithm to iteratively construct a sequence of incorrect geographic locations that form a path to the target. At each step, the attack generates subtle perturbations to the query image, guiding the geolocalization model toward the next location in this deceptive path. We show that our method is also strong in black-box settings, obtaining highly transferable attacks with less perceptible image artifacts.
  \keywords{Adversarial attack \and Image geolocalization \and Privacy}
\end{abstract}

\section{Introduction}

Recent advances in image-based geolocalization \cite{berton2022deep,vivanco2023geoclip, haas2024pigeon} pose pressing privacy and safety issues. Retrieval-based geolocalization methods  \cite{vivanco2023geoclip, jia2024g3} can accurately predict a location's country and even region from a single image. The public availability of such models allows malicious actors to invade user privacy and, in more critical cases, may even endanger their safety and well-being.
Modern image-based geolocalization models loosely fall into either retrieval or classification approaches. Retrieval-based methods rely on a large set of geolocated samples and some feature learning procedure. Classification-based methods generally partition the coordinate space hierarchically through classifiers. While this second approach has shown a slight edge in terms of precision, retrieval-based methods can be scaled up to incorporate more knowledge without relearning the feature space. Moreover, performing location recognition via retrieval, which compares a query embedding against a database of geotagged image prototypes, is formally equivalent to a classification task in which the feature vectors of the prototypes serve as the weight vectors for their respective location classes in a linear classifier.

In this paper, we address the privacy issues stemming from the widespread availability of geolocalization models through the lens of adversarial machine learning. 
We propose to protect user privacy by attacking image-based localization models. Adversarial methods are designed to produce a subtle perturbation that, when added to an image, induces erroneous behavior in the targeted model. Adversarial machine learning has been used in the past as a means of user privacy protection, for example in the field of text-based image editing~\cite{salman2023raising, shan2023glaze, liang2023mist, trippodo2025immunizing}. Here, perturbations are crafted to immunize user images from diffusion-based editing.
Despite the extensive body of work on adversarial machine learning~\cite{PGDpaper, szegedy2014, deepfool, autoattack, squarea, athalye2018synthesizing, elasticnet}, robustness studies specifically targeting modern geolocalization models remain virtually non-existent, with GeoShield~\cite{geoshield} representing the sole exception in the literature to date. 
We study the problem using different threat models: we first formulate a white-box approach, where we assume to have knowledge of the geolocalization model used by a malicious user; we then study the transferability of our approach in a black-box scenario, i.e. attacking a surrogate model, without having prior knowledge of the target geolocalization model.

\begin{figure}[!t]
    \centering
    \includegraphics[width=0.85\linewidth]{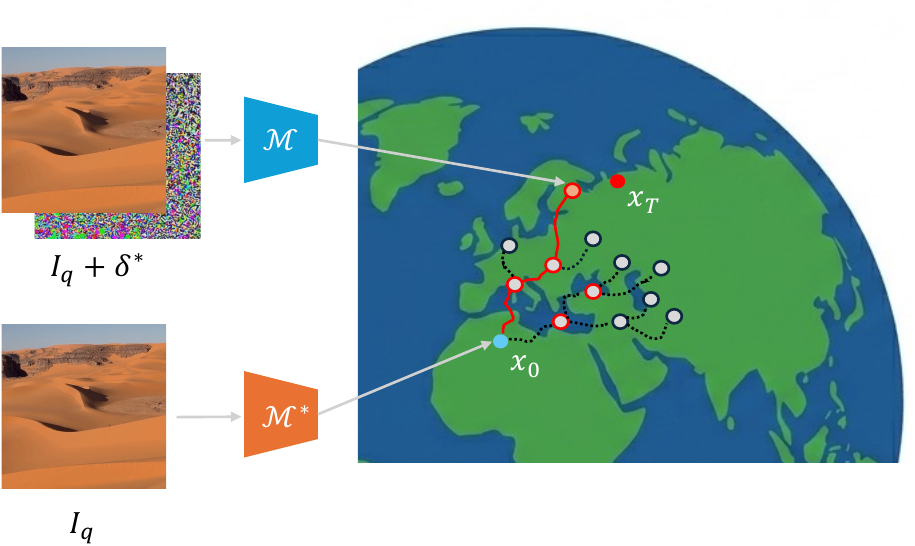}
\caption{RoadTrip Attack. Starting from an image $I_q$ correctly localized at $x_0$ by the victim model $\mathcal{M}^*$, the attack crafts an adversarial image $I_q + \delta^*$. By leveraging a surrogate model $\mathcal{M}$, the algorithm explores a ``beam'' of black-box adversarial solutions (red dots). These are selected as the optimal subset of iteratively sampled geographic neighbors (gray dots). The final optimal ``trip'' (highlighted in red) demonstrates the gradual shift from the ground truth to the intended adversarial target location $x_T$.}
    \label{fig:eyecatcher}
\end{figure}

First, we analyze the effectiveness of the state-of-the-art attacks such as Projected Gradient Descent (PGD) \cite{PGDpaper}, Carlini $\&$ Wagner \cite{carlini2017towards} and Fast Gradient Sign Method (FGSM) \cite{FGSMpaper} in disrupting retrieval models. We then use this analysis to craft an even more effective method that leverages the geographic nature of the geolocalization task. Attacks like PGD have a significant advantage over simpler non-iterative attack methods, such as FGSM: the use of multiple steps and random restarts. This iterative process suggests that PGD could be made even stronger if multiple optimization paths were explored simultaneously during the attack.

A naive solution might be to sample multiple random perturbations in the image feature space at each iteration to broaden the search. However, a key issue arises: we can make few assumptions about the structure of this visual space. Recent work has shown that the feature spaces of contrastively-trained models can contain discontinuities among embeddings of the same modality~\cite{mistretta2025cross, yi2024leveraging}. Furthermore, since perturbed images are out-of-distribution data, their behavior within this space is unpredictable.
Instead, we leverage the inherent geometrical structure of the problem. By performing targeted attacks, we can guide the model toward a set of intermediate geographic locations. We propose sampling these intermediate targets from a uniform distribution within a disk of a given radius around the target location.
A second issue arises regarding the radius of this sampling disk. To ensure convergence, we employ a simple solution: scaling the sampling radius by the Haversine distance between the current location and the final attack target. 
Finally, to keep our approach computationally tractable, we adapt beam search to this problem. At each iteration, a batch of random geographic targets is used to generate adversarial candidates, but we only continue the optimization process for a small subset (the ``beam'') of the most successful ones. This approach finds non-obvious sequences of optimization paths, resulting in more effective adversarial noise. Because the sequence of intermediate targets resembles stops on a road trip from the original location to the adversarial one, we name our method the RoadTrip Attack. A visual sketch of our approach is presented in Fig.~\ref{fig:eyecatcher}.

We evaluate our proposed attack by reporting white-box results on two popular geolocalization datasets. We also  show that the attack effectively transfers to two other state-of-the-art models in a black-box fashion.
Our results show that the RoadTrip Attack significantly outperforms standard adversarial attack baselines, as well as the VLM-based GeoShield attack, especially at extremely low perturbation budgets.
Our contribution is threefold:

\begin{itemize}

\item We address a pressing privacy and safety concern regarding modern image-based geolocalization methods leveraging adversarial machine learning.

\item We propose the RoadTrip Attack, a novel adversarial method specifically designed to exploit the inherent spatial structure of the geolocalization problem. RoadTrip improves over standard gradient-based methods by finding non-trivial optimization paths toward stronger attacks without requiring a higher noise budget.

\item {We empirically demonstrate that our specialized attack consistently and significantly outperforms strong, general-purpose baselines as well as geo\-localization-based attacks, requiring a small perturbation budget both in white-box and black-box settings.
}
\end{itemize}

\section{Related Work}
Since in this work we seek to protect privacy from image-based geolocalization through adversarial machine learning, in the following we first review recent  methods to estimate geographic coordinates from imagery and then we address adversarial machine learning literature.

\subsection{Visual Worldwide Geolocalization}
Visual worldwide geolocalization aims to predict the geographic coordinates of an image using only its visual content. 
Research in this area has evolved through three main methodological paradigms: classification-based~\cite{PlaNet,seo2018cplanet,muller2018geolocation,clark2023we,theiner2022interpretable,haas2024pigeon, pramanick2022world}, retrieval-based~\cite{berton2022deep, xu2023transvlad,hays_efros_2015, Oliva2006BuildingTG,vo2017revisiting,hausler2021patch,garg2021your,vivanco2023geoclip}, and generation-based approaches~\cite{jia2024g3, img2loc,jia2026georanker}. Some methods use a hybrid approach combining classification and regression~\cite{vo2017revisiting,kordopatis2021leveraging,haas2024pigeon,osv5m,izbicki2019exploiting}.

\paragraph{Classification-based methods}
The earliest approaches conceptualized geolocalization as a multi-class classification task, partitioning Earth's surface into discrete geographic cells. Pioneering work such as PlaNet~\cite{PlaNet} leverages structured grid systems like Google's S2 geometry to create hierarchical tessellations, where each cell represents a distinct class and the cell centroid serves as the estimated location~\cite{seo2018cplanet, muller2018geolocation}.

Researchers have explored diverse spatial partitioning methodologies from uniform regular grids\cite{PlaNet} to adaptive partitions based on data density~\cite{clark2023we}, semantic-driven divisions aligned with geographic features~\cite{theiner2022interpretable}, and administrative boundary-based partitions~\cite{haas2024pigeon, pramanick2022world}. 
Contemporary advances like PIGEON~\cite{haas2024pigeon} have elevated this paradigm by introducing semantically meaningful geocell construction with multi-task contrastive pretraining, achieving state-of-the-art performance while maintaining computational efficiency.

\paragraph{Retrieval-based methods}

Retrieval-based approaches reframe geolocalization as a nearest-neighbor search within high-dimensional feature spaces, assigning the coordinates of the most similar match as the predicted location~\cite{berton2022deep, xu2023transvlad}. Early implementations relied on handcrafted descriptors including color histograms~\cite{hays2008im2gps}, gist features~\cite{Oliva2006BuildingTG}, and SIFT descriptors with SVM classifiers~\cite{hays_efros_2015}. Deep learning architectures marked a transformative advancement~\cite{vo2017revisiting}, enabling end-to-end learning of robust visual representations with significantly enhanced matching accuracy.
GeoCLIP~\cite{vivanco2023geoclip} represents a notable breakthrough by introducing specialized location encoders using Random Fourier Features and Equal Earth projections, enabling direct image-to-GPS coordinate retrieval with hierarchical search capabilities.

\paragraph{Hybrid approaches}
Recognition of complementary strengths in classification and retrieval paradigms has motivated hybrid methodologies that strategically combine both approaches. These frameworks integrate discrete classification with continuous retrieval through ranking loss formulations~\cite{vo2017revisiting} and contrastive learning objectives~\cite{kordopatis2021leveraging}.
Notable implementations include classification-then-regression architectures that first predict coarse regions before applying fine-grained refinement~\cite{haas2024pigeon, osv5m}. Advanced hybrid approaches estimate probability distributions over geographic space using spherical Gaussian formulations~\cite{izbicki2019exploiting}, providing uncertainty quantification alongside location estimates.

\paragraph{Generation-based methods}
The most recent paradigm leverages large-scale multimodal models to generate location predictions through sophisticated reasoning processes. Systems like Img2Loc~\cite{img2loc}, G3~\cite{jia2024g3}, and GeoRanker~\cite{jia2026georanker} implement retrieval-augmented generation frameworks that synthesize visual features, GPS coordinates, and textual geographic descriptions within unified representational spaces.
Unlike conventional retrieval methods relying solely on visual similarity, generation-based approaches incorporate explicit geo-alignment mechanisms into image representations, enabling sophisticated reasoning for heterogeneous queries that may differ substantially from reference data~\cite{jia2024g3}. This paradigm demonstrates particular advantages when handling novel viewpoints, unusual lighting conditions, or rare geographic locations with limited training data.

Img2Loc~\cite{img2loc} leverages a two-step RAG workflow for global geolocalization. By performing an initial retrieval of visually analogous images, the model extracts spatial metadata to serve as in-context evidence. This context is then injected into the prompt, allowing the VLLM to produce more precise geographic estimates. G3~\cite{jia2024g3} utilizes a three-stage framework to improve geolocalization accuracy. First, it merges images, text, and coordinates into a single searchable format. Then, using an RAG-based approach, it identifies a diverse set of possible locations. Finally, it filters those candidates to ensure the most reliable result. GeoRanker~\cite{jia2026georanker} introduces a distance-aware ranking framework for worldwide image geolocalization that goes beyond simple visual similarity. It uses large vision-language models to jointly encode interactions between a query image and candidate locations, learning both absolute and relative geographic distances through a multi-order distance loss. This enables structured spatial reasoning over candidates and yields state-of-the-art performance on standard benchmarks.

\subsection{Adversarial Machine Learning}
Adversarial machine learning encompasses a variety of techniques designed to deliberately disrupt the performance of target models. These adversarial examples are crafted to be imperceptible to human observers while causing the model to produce significantly altered or incorrect outputs \cite{PGDpaper,FGSMpaper,szegedy2014, carlini2017towards}. 

The most common framework for generating these inputs is the additive threat model. Given an original image $x$, an adversarial counterpart is constructed as $x_{adv} = x + \delta$. To ensure the manipulation remains visually imperceptible, the perturbation $\delta$ is strictly constrained by a predefined budget $\varepsilon$, formulated as $||\delta||_{\infty} < \varepsilon$. 

Generating this perturbation typically involves maximizing the target model's loss, with established methods approaching this through varying optimization strategies. The Fast Gradient Sign Method (FGSM) \cite{FGSMpaper} generates the perturbation in a single step by moving in the direction of the gradient, while Projected Gradient Descent (PGD) \cite{PGDpaper} employs an iterative approach to maximize the loss while projecting back into the $\varepsilon$-budget constraint. In contrast, Carlini \& Wagner (C\&W) \cite{carlini2017towards} diverges from strict budget bounding by minimizing the additive perturbation $\delta$ alongside a soft-constraint on the adversarial loss; here, a constant $c$ acts as a hyperparameter to balance the priority of forcing misclassification against the goal of maintaining a minimal perturbation size.
Several adversarial generation methodologies rely on stochastic meta-heuris\-tics, such as differential evolution and evolutionary algorithms \cite{su2019one, nguyen2015deep}. To enhance transferability, FTQ-PSO \cite{li_optimizing_latent_variables} introduces a population-based heuristic within the latent space, employing particle swarm optimization in a gray-box setting. In contrast to these purely stochastic or swarm-based approaches, our method integrates gradient-based local optimization with a systematic, heuristic-guided beam search. This hybrid strategy allows for the simultaneous exploration of multiple high-potential search paths while ensuring each candidate is locally refined for maximum adversarial impact.

Despite the rapid evolution of adversarial techniques in general computer vision, the intersection of adversarial machine learning and image geolocalization remains underexplored. GeoShield\cite{geoshield}, which to our knowledge represents the only dedicated image-geolocalization-related adversarial method currently available in the literature, utilizes a feature disentanglement module to separate geographic from non-geographic information, alongside an exposure element identification module that pinpoints position-revealing regions within an image. By applying scale-adaptive adversarial perturbations to these specific areas, the model effectively misleads geolocation inference across various resolutions while maintaining the visual and semantic integrity of the original image.

\section{Method}
Our method leverages the intrinsic geometric structure of the image-based geolocalization problem. Similarly to other iterative \cite{kurakin2018adversarial, PGDpaper} adversarial attacks, we create a sequence of image perturbations to induce a distortion in the feature space in order to make localization inaccurate. Our method is based on alternating beam search and projected gradient descent \cite{PGDpaper} optimization. 
The main idea of attacking a geolocalization method is to target a location $x_T$ that differs from the ground truth location $x_0$ and alter the image representation so that features are incorrectly placed close to the requested target. 

\subsection{Attacking Retrieval Based Localization}
Our approach utilizes projected gradient descent \cite{PGDpaper} to attack retrieval-based geolocalizers. In this framework, the model leverages a gallery of pre-computed embeddings $\mathcal{D}: \{\mathbf{e}_0, \ldots, \mathbf{e}_k\}$. Depending on the architecture, these represent either a collection of geo-located reference images or a set of geographic coordinate embeddings. The system processes a query image $I_q$ through an encoder to produce a query embedding. It then computes similarity scores $s_i$ between the query and each gallery entry:
\begin{equation}
    s_i = \langle \text{enc}(I_q), \mathbf{e}_i\rangle
    \label{eq:sim}
\end{equation}

The final geographic prediction is derived from the coordinates associated with the gallery embeddings that yield the highest similarity scores. Our method seeks to perturb $I_q$ such that the similarity scores are redirected away from the ground-truth region toward distant, incorrect entries in the gallery.

\nn{
Therefore, for a targeted attack against such methods, we minimize the cross-entropy loss with respect to the target embedding:

\begin{equation}
    \mathcal{L}_{\mathrm{CE}}\left(\text{enc}(I_q), \mathbf{e}_T\right) = - \sum_{i=1}^{C} q_i \log p_i,
\end{equation}

where $p_i$ are softmax outputs of scores $s_i$ and $q_i = \mathbf{1}_{\{i=T\}}$. Here, $C$ is the number of gallery images, and $T$ is the index of the target embedding.

Each internal iteration of our method will update the image $I_q$ according to:

\begin{equation}
    I_q^{(t+1)} = \Pi_{\mathcal{B}_\epsilon(I_q)} \left( I_q^{(t)} - \alpha \, \mathrm{sign}\left( \nabla_{I_q^{(t)}} \mathcal{L}_{\mathrm{CE}}\left(\text{enc}\left(I_q^{(t)}\right), \mathbf{e}_T\right) \right) \right)
\end{equation}

where $\alpha$ is the iteration step, $\epsilon$ is the attack budget, and $\Pi_{\mathcal{B}_\epsilon(I_q)}$ is the projection operator clipping the adversarial image in the $l_\infty$ ball centered at the original query image $I_q$ with radius $\epsilon$.
}

\begin{algorithm}[t]

\begin{algorithmic}[1]
\REQUIRE $I_q$,$x_q$, $x_T$, $\epsilon$, $\alpha$, $\mathcal{L(\cdot)}$, $\mathcal{G(\cdot)}$, $\eta$, $N$, $J$, $K$
\STATE $\delta_1 \sim \mathcal{U}_{B}$ \textit{//Initialize perturbation} 
\STATE $\mathcal{S}_1 \gets \{(x_q, \delta_1)\}$ \textit{//Initialize source set}
\FOR{$i = 1$ to $N$}
    \STATE  $\mathcal{S}^*_{i+1}\gets \emptyset$  \textit{//Initialize next source candidates set}
    \FOR { ($x_{m,i},\delta_{m,i}) \in \mathcal{S}_i$}
 
 \STATE $\mathrm{R} \gets \eta \mathcal{G} (x_{m,i}, x_T)$  \textit{//Estimate sampling radius}
 \STATE $\mathcal{T}_i \gets \{\tau_{m}^j | \tau_{m}^j \sim \mathcal{U}_{D_\mathrm{R}(x_{T,i})}, ~ \forall j \in [1 \ldots J] \}$ \textit{//Populate target set}
 \FOR {$\tau_{m}^j \in \mathcal{T}_i$}
 \STATE $\delta^j_{m,i+1} \gets \mathrm{P}(I_q, \tau_m^j, \delta_{m,i} ,\epsilon, \alpha, \mathcal{L})$ \textit{//Update prev. noises $\delta_{m,i}$ for targets $\tau_m^j$}
 \STATE $x^j_{m,i+1} \gets \mathcal{M} (I_q+\delta^j_{m,i+1})$
 \STATE $\mathcal{S}^*_{i+1} \gets \mathcal{S}^*_{i+1} \cup \{(x^j_{m,i+1}, \delta^j_{m,i+1})\}$
 
 \ENDFOR
    \ENDFOR
    \STATE $\mathcal{S}_{i+1} \gets \mathrm{top}_K (\mathcal{S}^*_{i+1}, x_T, \mathcal{G})$ 
\ENDFOR
\STATE $ (x^*, \delta^*) \gets \arg\!\min_{(x,\delta)}  \mathcal{G}(x, x_T)~\mathrm{s.t.} ~ (x,\delta) \in \mathcal{S}_N$
\RETURN $\delta^*$

\end{algorithmic}
\caption{RoadTrip Attack.\label{algo:roadtrip}}
\end{algorithm}

\subsection{RoadTrip Attack}

The optimization landscape for adversarial attacks on geolocalization models is challenging. Visually similar features, such as comparable architecture or vegetation, can exist in geographically distant locations, creating a highly complex and non-convex feature space. A standard gradient-based optimization like PGD \cite{PGDpaper} follows a direct, greedy path toward the target location. This approach is susceptible to getting trapped in poor local minima, where the model's output is closer to the target but the required perturbation is unnecessarily large or sub-optimal.

To overcome this, the RoadTrip Attack reframes the problem. Instead of a direct attack toward the adversarial target, we conceptualize the attack as a journey. The intuition is that finding a sequence of easier, intermediate steps can lead to a more effective final perturbation. Our approach explores multiple branching paths in parallel, using beam search to prune unpromising routes. This method acts as a form of ``geographic annealing'', allowing the optimization to navigate the complex feature space more effectively and discover non-obvious, stronger adversarial solutions that a direct path would miss.

The main idea of the RoadTrip attack is to avoid a direct path toward the target but instead iteratively sample intermediate targets in a disk around the final target and leverage beam search to pick the most promising top-k candidates for the next step. Unlike PGD, which optimizes toward a fixed target, our method concurrently builds multiple paths toward the final destination. This iterative process of generating and pruning candidate solutions allows the attack to discover more effective adversarial paths. A detailed, step-by-step description of our method is provided in Alg. \ref{algo:roadtrip} and is explained in the remainder of this section.

In the following, we measure distances between coordinates on the earth using the Haversine distance
 \begin{equation}
\mathcal{G}= 2R \sin^{-1}\!\left(
  \sqrt{ \operatorname{h}(\phi_2-\phi_1)
+ \cos(\phi_1)\cos(\phi_2)\,\operatorname{h}(\lambda_2-\lambda_1)}\right)
\end{equation}

 which measures the shortest path between coordinates $(\phi_1, \lambda_1$) and $(\phi_2, \lambda_2$), expressed in latitude $\phi_i$ and longitude $\lambda_i$, given the average Earth radius $R$, and where $\operatorname{h}(\theta) = \sin^2\!\left(\tfrac{\theta}{2}\right)$.

Our method iteratively grows and prunes a set of candidate starting points $\mathcal{S}_i$ and perturbations $\mathcal{N}_i$.  The state of our algorithm is retained by a set of image perturbation values $\mathcal{N}_i:\{\delta_{1,i}\ldots \delta_{K,i}\}$ and a set of geographic coordinates $\mathcal{S}_i:\{ x_{1,i}\ldots x_{K,i}\}$.

At any given iteration $i$, for each starting point $x_{m,i}\in\mathcal{S}_i$ we uniformly sample a set of target coordinates $\tau^j_m$ in a disk $D_R(x_T)$ 
of radius R, centered in $x_T$. 
For each of the sampled locations $\tau^j_m$, we create a perturbation via projected-gradient optimization, aiming at altering the image representation to make it geographically closer to location $\tau^j_m$, according to some loss $\mathcal{L}$, which in its vanilla form is a cross-entropy loss $\mathcal{L}_{CE}$.

Let \begin{equation}
    \delta_{m,i+1} \gets \mathrm{P}(I_q, \tau_m^j, \delta_{m,i} ,\epsilon, \mathcal{L})
\end{equation}

be the projected gradient descent attack on query image $I_q$ directed toward location $\tau_m^j$ which, at convergence, returns the best perturbation $\delta_{m,i+1}$. Such a perturbation, when applied to $I_q$, leads the model to yield the corresponding location $x_{m,i+1}$.

We initialize the noise $\delta_{m,i}$ by uniformly sampling in $B(\epsilon)$ where
\begin{equation}
B(\epsilon) = \left\{ \delta \in \mathbb{R}^n : \|\delta\|_{\infty} \leq \epsilon \right\},
\end{equation}

We name $\mathcal{T}_{i+1}$ the set of all attacks converged to solutions $x_{m, i+1}$, and pick the top-K according to $\mathcal{G}$. Since we are targeting $x_T$ we pick the $K$ closest solutions to $x_T$ as the subsequent iteration of starting points $\mathcal{S}_{i+1}$. 

As in beam search, we iteratively drop less promising solutions and move toward our target set of candidates. After $N$ steps, we pick the best perturbation $\delta_{i^*}$ as the one corresponding to $x_{i^*} \in \mathcal{S}_N$, i.e. the closest solution to $x_T$. An important implementation detail is the choice of the sampling radius R. A fixed radius presents a trade-off: a large radius would be inefficient for fine-grained adjustments when the solution is already close to the target, while a small radius would make the trip exceedingly slow and computationally expensive when far away.
To resolve this, we employ an adaptive radius strategy. As outlined in Alg.~\ref{algo:roadtrip}, the radius R is dynamically scaled in proportion to the Haversine distance between the current candidate location $x_{m,i}$ and the final adversarial target $x_{T}$.
When the current solution is geographically distant from the final target, the algorithm uses a larger radius to explore wider jumps, rapidly closing the distance.
Conversely, as the attack path converges toward the target, the sampling radius automatically shrinks. This narrowing focus allows for a more fine-grained search, enabling the precise optimization required to reach the intended location. In our experiments, we set the scaling factor $\eta$ to 0.5.

The target location $x_T$ is sampled randomly from a gallery of GPS coordinates under the condition that the Haversine distance between $x_0$ and $x_T$ is more than 2500km.
Other than that, the GPS gallery is used only during noise optimization to compute cross-entropy loss (which is inherent to retrieval-based models). Thus, RoadTrip Attack does not require access to the full database for intermediate target selection.

In the black-box setting, we use GeoCLIP \cite{vivanco2023geoclip} as a surrogate for the attacked models. Notably, this approach requires access to neither the victim’s internal architecture nor its gallery.

\section{Experiments}
\subsection{Datasets and Metrics}
We evaluate our proposed attack on two geolocalization benchmarks, Im2GPS3k and YFCC4k~\cite{vo2017revisiting}.  IM2GPS3k consists of around 3000 images from Im2GPS \cite{hays2008im2gps}. Note that this differs from the original Im2GPS test set \cite{hays2008im2gps}.  YFCC4k comprises around 4000 random images from the YFCC100m \cite{thomee2016yfcc100m}
dataset. YFCC100m was not collected specifically for geolocalization but for generic computer vision tasks. Therefore, its distribution is different from Im2GPS and it is considered more challenging.
To evaluate the effectiveness of our attack, we use three different metrics. Ground Truth Accuracy $ACC_{GT}$ is the primary metric to assess the success of the attack. It measures the percentage of attacked images that are still localized within a distance threshold of their original true location. For a successful attack, this value should be as low as possible, demonstrating the model has been successfully fooled.
Target Accuracy $ACC_{Target}$ measures the percentage of attacked images whose predicted location falls within a certain Haversine distance threshold from the chosen adversarial target location $x_T$. A higher $ACC_{Target}$ value indicates a more successful targeted attack.
We evaluate the attack for different $ACC_{GT}$ and $ACC_{Target}$ distance thresholds, namely 1km, 25km, 200km, 750km, and 2500km, computed with the Haversine distance.

We vary the budget for all attacks, setting $\epsilon=~\{2/255, 4/255\}$ for PGD, RTA and FGSM and $c = \{0.1, 1\}$ for Carlini-Wagner. Considering the high target accuracy reached by RTA at 1km we do not test larger budgets. 

\subsection{Implementation Details}
All experiments have been run on a RTX 5000 Blackwell.
For the models, we used the official codebases of GeoCLIP \cite{vivanco2023geoclip},  G3 \cite{jia2024g3}, and a custom implementation of Img2Loc\cite{img2loc} based on the one used as baseline by Jia et al. \cite{jia2024g3}; this was necessary since the official codebase of Img2Loc is not complete.

\subsection{White-Box Results}

We target a state-of-the-art geolocalization model, GeoCLIP \cite{vivanco2023geoclip},
a retrieval method leveraging contrastively learned representations of images and coordinates.
We benchmark our approach against widely used  adversarial baselines, namely FGSM \cite{FGSMpaper}, PGD \cite{PGDpaper}, and CW \cite{carlini2017towards} on the Im2GPS3k and YFCC4k datasets~\cite{vo2017revisiting}, to ensure a comprehensive robustness evaluation.

\begin{table*}[t]
\centering
\resizebox{\textwidth}{!}{
\begin{tabular}{l|c|cH|ccccc|ccccc}
 \multicolumn{4}{c}{} & \multicolumn{5}{c}{ACC$_{GT}$ ($\downarrow$)} & \multicolumn{5}{c}{ACC$_{Target}$ ($\uparrow$)}\\ \hline
 & $\varepsilon$ & LPIPS ($\downarrow$) & PSNR ($\uparrow$) & @1km & @25km & @200km & @750km & @2500km & @1km & @25km & @200km & @750km & @2500km \\ \hline
CW & c = 0.1 & 0.031 & 47.613 & 0.43\% & 1.63\% & 2.80\% & 5.44\% & 10.88\% & 53.79\% & 55.59\% & 59.26\% & 63.90\% & 74.74\% \\
CW & c = 1.0 & 0.079 & 42.027 & 0.10\% & 0.43\% & 0.63\% & 1.40\% & 3.97\% & 81.95\% & 83.35\% & 84.35\% & 86.55\% & 90.52\% \\
\hline
FGSM & \multirow{3}{*}{2/255} & 0.007 &	51.157	& 6.64\% &	15.98\% & 23.69\% &	37.30\%	& 55.39\% &	0.33\%	& 0.33\% & 0.53\% & 1.77\%	& 9.41\% \\

PGD &  & 0.007 & 51.225 & 0.40\% & 1.03\% & 2.10\% & 4.34\% & 8.34\% & 71.81\% & 73.97\% & 76.24\% & 79.71\% & 85.82\% \\
RTA & &{0.027} & \textbf{47.001} & \textbf{0.03\%} & \textbf{0.03\%} & \textbf{0.03\%} & \textbf{0.10\%} & \textbf{1.03\%} & \textbf{93.43\%} & \textbf{94.99\%} & \textbf{95.70\%} &\textbf{96.96\%} & \textbf{98.50\%} \\ \hline

FGSM & \multirow{3}{*}{4/255} & 0.042 & 44.186 & 4.57\% & 12.08\% & 18.82\% & 30.93\% & 49.15\% & 0.40\% & 0.40\% & 0.93\% & 2.50\% & 11.64\% \\
PGD &  & 0.044 & 44.403 & \textbf{0.00\%} & 0.07\% & 0.23\% & 0.33\% & 1.27\% & 96.63\% & 98.13\% & 98.47\% & 98.67\% & 99.03\% \\
RTA & & 0.059 & 43.211 & \textbf{0.00\%} & \textbf{0.00\%} & \textbf{0.00\%} & \textbf{0.03\%} & \textbf{0.73\%} & \textbf{98.93\%} & \textbf{99.73\%} &\textbf{ 99.83\%} & \textbf{99.87\%} & \textbf{99.93\%} \\
\hline
\end{tabular}
}
 \caption{{Performance of RTA attacking GeoCLIP \cite{vivanco2023geoclip} on Im2GPS3k \cite{vo2017revisiting}.}}
\label{tab:geoclip_im2gps}
\end{table*}

\begin{table*}[t]
\centering
\resizebox{\textwidth}{!}{
\begin{tabular}{l|c|cH|ccccc|ccccc}
 \multicolumn{4}{c}{} & \multicolumn{5}{c}{ACC$_{GT}$ ($\downarrow$)} & \multicolumn{5}{c}{ACC$_{Target}$ ($\uparrow$)}\\ \hline
 & $\varepsilon$ & LPIPS ($\downarrow$) & PSNR ($\uparrow$) & @1km & @25km & @200km & @750km & @2500km & @1km & @25km & @200km & @750km & @2500km \\ \hline
CW & c = 0.1 & 0.022 & 48.793 & 0.35\% & 0.77\% & 1.90\% & 4.41\% & 10.10\% & 47.38\% & 49.27\% & 52.40\% & 57.50\% & 69.49\% \\
CW & c = 1.0 & 0.054  & 43.464 & 0.13\% & 0.26\% & 0.49\% & 1.41\% & 4.45\% & 76.85\% & 78.62\% & 80.42\% & 82.43\% & 87.57\% \\
\hline
FGSM & \multirow{3}{*}{2/255} & 0.008 & 51.214 & 4.14\% & 8.42\% & 15.70\% & 30.91\% & 49.80\% & 0.24\% & 0.24\% & 0.44\% & 1.74\% & 9.55\% \\

PGD &  & 0.009 & 51.225 & 0.15\% & 0.37\% & 0.82\% & 2.07\% & 4.23\% & 62.30\% & 64.30\% & 67.12\% & 71.62\% & 80.44\% \\
RTA & & 0.031 & 46.692 & \textbf{0.04\%} & \textbf{0.07\%} & \textbf{0.09\%} & \textbf{0.24\%} & \textbf{1.19\%} & \textbf{88.60\%} & \textbf{90.37\%} & \textbf{92.15\%} & \textbf{94.44\%} & \textbf{97.49\%} \\
\hline

FGSM & \multirow{3}{*}{4/255} & 0.041 & 44.278 & 3.28\% & 6.33\% & 11.73\% & 25.62\% & 44.40\% & 0.26\% & 0.26\% & 0.57\% & 2.14\% & 11.44\% \\
PGD &  &  0.044 & 44.485 & 0.04\% & 0.13\% & 0.22\% & 0.42\% & 1.34\% & 94.80\% & 96.49\% & 97.00\% & 97.57\% & 98.48\% \\
RTA & & 0.062 & 42.960 & \textbf{0.02\%} & \textbf{0.02\%} & \textbf{0.02\%} & \textbf{0.02\%} & \textbf{0.68\%} & \textbf{97.82\%} & \textbf{99.01\%} &\textbf{ 99.21\%} & \textbf{99.51\%} & \textbf{99.82\%} \\
 \hline
\end{tabular}
}
\caption{{Performance of RTA attacking GeoCLIP \cite{vivanco2023geoclip} on YFCC4k \cite{vo2017revisiting}.}}
\label{tab:geoclip_yfcc}
\end{table*}

Our experimental results, summarized in Tab. \ref{tab:geoclip_im2gps} and Tab. \ref{tab:geoclip_yfcc}, demonstrate that the RoadTrip Attack consistently outperforms all baselines across all models and datasets.  Among the baselines, PGD perform best. FGSM evidently suffers from the single-step attack and does not allow sufficiently complex perturbations to be crafted. CW has better performance with respect to FGSM, but still underperforms versus PGD.
The benefits of RTA emerge more prominently in low-budget settings ($\epsilon=2/255$). For instance, on the Im2GPS3k dataset (Tab. \ref{tab:geoclip_im2gps}), with a minimal perturbation budget $\epsilon=2/255$, the standard PGD attack only reduces the $ACC_{GT}@2500km$ to 8.34\%, whereas RTA in its vanilla formulation lowers it to 1.03\%. At the same time, RTA achieves a much higher target success rate with the same budget, increasing $ACC_{Target}@1km$ from 71.81\% to 93.43\%. This pattern is consistent across datasets.

\ls{To highlight the contribution of the geographical framing of the RTA problem, we compare our results with several PGD-based variants under the same perturbation constraint, $\varepsilon=2/255$ (Tab.~\ref{tab:comparison_pgd_versions_rta}).}\nn{\textit{Multi-start} runs five PGD optimizations in parallel and selects the best adversarial example. \textit{Random restart} randomly perturbs the adversarial noise during optimization, introducing stochastic exploration. \textit{Interpolated} uses five deterministic intermediate targets obtained by interpolating between the source and target locations. RTA consistently achieves lower ACC$_{GT}$ and higher ACC$_{Target}$ than all variants, indicating that its improvement is not only due to a larger search budget or random exploration, but to the adaptive selection of intermediate geographic targets.}

\begin{table*}[t]
\centering
\resizebox{\textwidth}{!}{
\begin{tabular}{l|ccccc|ccccc}
 \multicolumn{1}{c}{} & \multicolumn{5}{c}{ACC$_{GT}$ ($\downarrow$)} & \multicolumn{5}{c}{ACC$_{Target}$ ($\uparrow$)}\\ \hline
  & @1km & @25km & @200km & @750km & @2500km & @1km & @25km & @200km & @750km & @2500km \\ \hline
 Clean & 14.11\% & 34.47\% & 50.65\% & 69.67\% & 83.82\% & -- & -- & -- & -- & -- \\ \hline
 PGD & 0.40\% & 1.03\% & 2.10\% & 4.34\% & 8.34\% & 71.81\% & 73.97\% & 76.24\% & 79.71\% & 85.82\% \\
 Multi start & 0.90\% & 0.90\% & 2.70\% & 3.60\% & 5.41\% & 72.07\% & 76.58\% & 79.28\% & 85.59\% & 89.19\% \\
 Random restart & 0.45\% & 0.90\% & 2.52\% & 4.41\% & 8.55\% & 64.27\% & 66.70\% & 69.94\% & 75.16\% & 82.90\% \\
 Interpolated & 0.99\% & 1.89\% & 3.60\% & 6.84\% & 13.77\% & 32.13\% & 33.57\% & 39.51\% & 47.79\% & 68.32\% \\
 RTA & \textbf{0.03\%} & \textbf{0.03\%} & \textbf{0.03\%} & \textbf{0.10\%} & \textbf{1.03\%} & \textbf{93.43\%} & \textbf{94.99\%} & \textbf{95.70\%} & \textbf{96.96\%} & \textbf{98.50\%} \\ \hline
\end{tabular}
}
\caption{{Comparison between PGD variants and RTA attacking GeoCLIP \cite{vivanco2023geoclip} on Im2GPS3k \cite{vo2017revisiting}.}}
\label{tab:comparison_pgd_versions_rta}
\end{table*}

\begin{table}[t]
\centering
\resizebox{0.9\textwidth}{!}{%
\begin{tabular}{l|cH|ccccc}
\multicolumn{3}{c}{} & \multicolumn{5}{c}{ACC$_{GT}$ ($\downarrow$)}\\
\toprule
\textbf{Method~~} & $\varepsilon$                 & LPIPS          & ~~~~@1km~~~~            & ~~~@25km~~~           & ~~~@200km~~~         & ~~~@750km~~~           & ~~~@2500km~~~ \\ \midrule
GeoShield       & \multirow{2}{*}{2/255}  & 0.047          & 8.54\%          & 24.42\%         & 33.73\%         & 47.45\%          & 63.43\%                     \\
RTA &  & \textbf{0.039} & \textbf{2.77\%} & \textbf{7.61\%} & \textbf{10.14\%} & \textbf{16.08\%} & \textbf{27.86\%} \\ \midrule
GeoShield       & \multirow{2}{*}{4/255}  & 0.097          & 7.57\%          & 19.99\%         & 26.63\%         & 38.14\%          & 55.52\%                     \\
RTA             &                         & \textbf{0.069} & \textbf{2.20\%} & \textbf{6.14\%} & \textbf{8.31\%} & \textbf{13.78\%} & \textbf{24.52\%}            \\ \midrule
GeoShield       & \multirow{2}{*}{8/255}  & 0.183          & 4.87\%          & 13.41\%         & 17.22\%         & 27.29\%          & 44.24\%                     \\
RTA             &                         & \textbf{0.108} & \textbf{2.20\%} & \textbf{5.74\%} & \textbf{7.47\%} & \textbf{12.61\%} & \textbf{23.69\%}            \\ \midrule
GeoShield       &  & 0.305          & 2.74\%          & 8.31\%          & 10.98\%         & 17.62\%          & 32.10\%                     \\ 
RTA             &    \multirow{-2}{*}{16/255}                     & \textbf{0.174} & \textbf{1.80\%} & \textbf{4.94\%} & \textbf{6.24\%} & \textbf{11.04\%} & \textbf{21.19\%}            \\ \bottomrule
\end{tabular}%
}
\caption{Black-box transferability from GeoCLIP to Img2Loc on Im2GPS3k.}
\label{tab:transfer_img2loc}
\end{table}


\begin{table}[t]
\centering
\resizebox{0.9\textwidth}{!}{%
\begin{tabular}{l|cH|ccccc}
\multicolumn{3}{c}{} & \multicolumn{5}{c}{ACC$_{GT}$ ($\downarrow$)}\\
\toprule
\textbf{Method~~} & $\varepsilon$                 & LPIPS          & ~~~~@1km~~~~            & ~~~@25km~~~           & ~~~@200km~~~         & ~~~@750km~~~           & ~~~@2500km~~~ \\ \midrule
GeoShield       & \multirow{2}{*}{2/255}  & 0.047          & 12.38\%         & 32.97\%         & 44.98\%         & 62.46\%         & 78.14\%                     \\
RTA             &                         & \textbf{0.039} & \textbf{1.17\%} & \textbf{3.04\%} & \textbf{4.54\%} & \textbf{8.07\%} & \textbf{19.12\%}            \\ \midrule
GeoShield       & \multirow{2}{*}{4/255}  & 0.097          & 9.71\%          & 26.43\%         & 36.07\%         & 50.58\%         & 68.67\%                     \\
RTA             &                         & \textbf{0.069} & \textbf{0.80\%} & \textbf{2.34\%} & \textbf{3.17\%} & \textbf{6.11\%} & \textbf{15.42\%}            \\ \midrule
GeoShield       & \multirow{2}{*}{8/255}  & 0.183          & 6.67\%          & 17.08\%         & 22.96\%         & 35.27\%         & 54.99\%                     \\
RTA             &                         & \textbf{0.108} & \textbf{0.77\%} & \textbf{2.67\%} & \textbf{3.67\%} & \textbf{6.67\%} & \textbf{16.15\%}            \\ \midrule
GeoShield       & \multirow{2}{*}{16/255} & 0.305          & 3.74\%          & 10.71\%         & 14.58\%         & 24.42\%         & 45.01\%                     \\
RTA             &                         & \textbf{0.174} & \textbf{0.73\%} & \textbf{2.27\%} & \textbf{3.50\%} & \textbf{6.47\%} & \textbf{16.08\%}            \\ \bottomrule
\end{tabular}%
}
\caption{Black-box transferability from GeoCLIP to G3 on Im2GPS3k.}
\label{tab:transfer_g3}
\end{table}

\subsection{Black-Box Results}

\begin{figure*}[t]
    \centering
    \resizebox{.97\linewidth}{!}{%
    \setlength{\tabcolsep}{4pt}%

        \begin{tabular}{c l l l l}
            & ~~~~\textbf{$\varepsilon=2/255$} & ~~~~\textbf{$\varepsilon=4/255$} & ~~~~\textbf{$\varepsilon=8/255$} & ~~~~\textbf{$\varepsilon=16/255$} \\

            \raisebox{0.6\height}{\rotatebox{90}{Geoshield\cite{geoshield}}} &
\begin{tikzpicture}[
    spy using outlines={circle, red, line width=1pt, magnification=3, size=2.5cm, connect spies}
]
    \node {\includegraphics[width=0.18\textwidth]{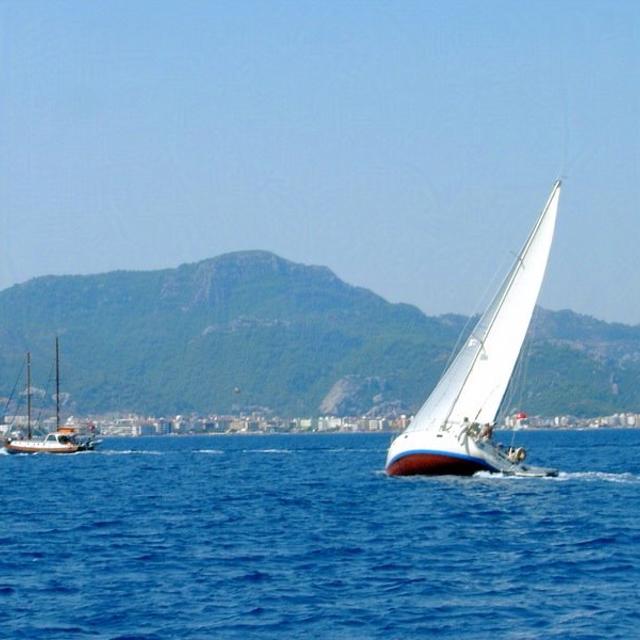}};
    \spy on (-0.6, 0.6) in node [right] at (0.1, -0.7);
\end{tikzpicture} &
            \begin{tikzpicture}[
    spy using outlines={circle, red, line width=1pt, magnification=3, size=2.5cm, connect spies}
]
    \node {\includegraphics[width=0.18\textwidth]{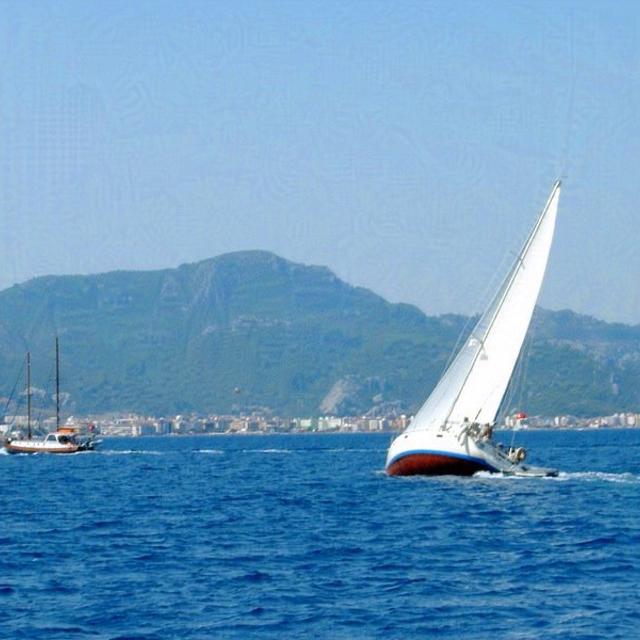}};
    \spy on (-0.6, 0.6) in node [right] at (0.1, -0.7);
\end{tikzpicture} &
            \begin{tikzpicture}[
    spy using outlines={circle, red, line width=1pt, magnification=3, size=2.5cm, connect spies}
]
    \node {\includegraphics[width=0.18\textwidth]{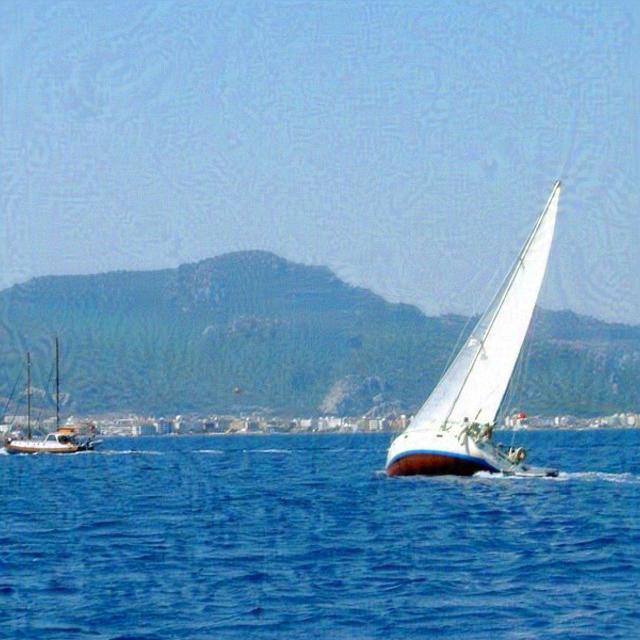}};
    \spy on (-0.6, 0.6) in node [right] at (0.1, -0.7);
\end{tikzpicture} &
            \begin{tikzpicture}[
    spy using outlines={circle, red, line width=1pt, magnification=3, size=2.5cm, connect spies}
]
    \node {\includegraphics[width=0.18\textwidth]{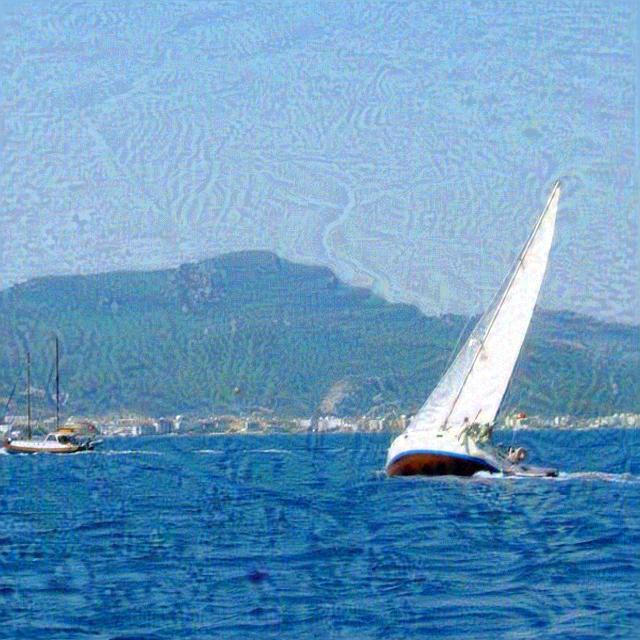}};
    \spy on (-0.6, 0.6) in node [right] at (0.1, -0.7);
\end{tikzpicture} \\

            \raisebox{0.6\height}{\rotatebox{90}{~~~~~~~RTA}} &
\begin{tikzpicture}[
    spy using outlines={circle, red, line width=1pt, magnification=3, size=2.5cm, connect spies}
]
    \node {\includegraphics[width=0.18\textwidth]{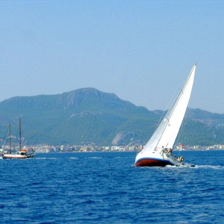}};
    \spy on (-0.6, 0.6) in node [right] at (0.1, -0.7);
\end{tikzpicture} &
            \begin{tikzpicture}[
    spy using outlines={circle, red, line width=1pt, magnification=3, size=2.5cm, connect spies}
]
    \node {\includegraphics[width=0.18\textwidth]{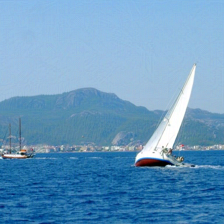}};
    \spy on (-0.6, 0.6) in node [right] at (0.1, -0.7);
\end{tikzpicture} &
            \begin{tikzpicture}[
    spy using outlines={circle, red, line width=1pt, magnification=3, size=2.5cm, connect spies}
]
    \node {\includegraphics[width=0.18\textwidth]{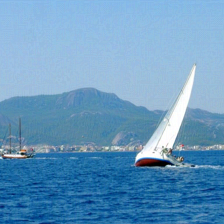}};
    \spy on (-0.6, 0.6) in node [right] at (0.1, -0.7);
\end{tikzpicture} &
            \begin{tikzpicture}[
    spy using outlines={circle, red, line width=1pt, magnification=3, size=2.5cm, connect spies}
]
    \node {\includegraphics[width=0.18\textwidth]{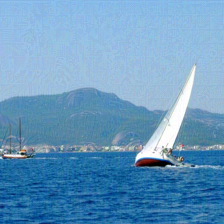}};
    \spy on (-0.6, 0.6) in node [right] at (0.1, -0.7);
\end{tikzpicture} \\

        \end{tabular}%
    }
    \caption{Visual comparison of adversarial noise for four budgets $\varepsilon = 2/255$, $4/255$, $8/255$, $16/255$ and two methods (Geoshield and ours). The magnified area highlights the fine-grained perturbation patterns for the attack. Best viewed in color on a screen.}
    \label{fig:ours_geoshield_comparison}
\end{figure*}

To evaluate our method under a more realistic threat model, we tested RTA against image geolocalization models different from the one used for the attack. We focus on Img2Loc \cite{img2loc} and G3 \cite{jia2024g3}, motivated by their state-of-the-art performance.
We compare RTA to GeoShield \cite{geoshield}, which to the best of our knowledge is the only available image geolocalization adversarial method existing in the literature. Note that the gallery used in the retrieval step of both these methods, MP16~\cite{larson2017benchmarking}, differs from the one on which we learn the adversarial noise, which is more than 40$\times$ smaller. 
In Tab. \ref{tab:transfer_img2loc} and Tab. \ref{tab:transfer_g3} we report ground truth accuracies for the attacks. We do not report target accuracy as GeoShield is untargeted. From the tables, it is clear that our method produces more disruptive adversarial perturbations. It is also very important to note that the adversarial noise added by RTA better preserves visual fidelity.
To measure this, we compute the Learned Perceptual Image Patch Similarity (LPIPS) \cite{zhang2018unreasonable}, which measures the visual quality and subtlety of the generated perturbations. LPIPS calculates the perceptual distance between the original and the perturbed image. Lower LPIPS values signify that the adversarial image is visually similar to the original, making the attack imperceptible to a human observer.
As shown in Tab. \ref{tab:lpips}, RTA alters the visual quality of the image in a much more subtle way compared to GeoShield, consistently across perturbation budgets. This difference is especially noticeable for high perturbation budgets, e.g. the LPIPS for GeoShield for $\varepsilon=16/255$ is almost twice the LPIPS for RTA. This behavior can be clearly observed qualitatively in Fig. \ref{fig:ours_geoshield_comparison}, where samples of attacked images are reported for different budgets. The adversarial noise for GeoShield is clearly visible to the human eye.

\begin{table}[t]
    \centering
    \begin{tabular}{l|c|c|c||c|c|c}
    \toprule
         \textbf{Method} & $\varepsilon$ & LPIPS ($\downarrow$) & PSNR ($\uparrow$) & $\varepsilon$& LPIPS ($\downarrow$) & PSNR ($\uparrow$)\\
         \midrule
         GeoShield & & 0.047 & 35.26 & & 0.183 & 32.96 \\
        RTA & \multirow{-2}{*}{$2/255$} & 
        \textbf{0.027} & \textbf{41.42} &\multirow{-2}{*}{$8/255$} & \textbf{0.108} & \textbf{36.07}\\
         \midrule
         GeoShield & & 0.097 & 34.60 & & 0.305 & 29.94\\
        RTA & \multirow{-2}{*}{$4/255$} &
        \textbf{0.059} & \textbf{38.57} & \multirow{-2}{*}{$16/255$}&  \textbf{0.174} & \textbf{32.90}\\
        
        \bottomrule
         
    \end{tabular}
    \caption{RTA adds less visible noise than GeoShield as highlighted by the lower LPIPS \nn{and the higher PSNR.}}
    \label{tab:lpips}
\end{table}

\subsection{Runtime} 
\begin{figure}[t]
     \centering
     
     \begin{subfigure}[b]{0.49\textwidth}
         \centering

        \begin{tikzpicture}
        \begin{axis}[
            width=\columnwidth, 
            xlabel={Beam Size},
            ylabel={Time (s/img)},
            xmin=0.5, xmax=4.5,
            ymin=1.0, ymax=4.0,
            xtick={1, 2, 4},
            ytick={1.0, 1.5, 2.0, 2.5, 3.0, 3.5},
            legend pos=south east,
            ymajorgrids=true,
            xmajorgrids=true,
            grid style=dashed,
            tick label style={/pgf/number format/fixed},
            label style={font=\scriptsize},
            ticklabel style={font=\scriptsize},
            title style={font=\small},
            legend style={font=\scriptsize},
            height=4.5cm
        ]
        
        \addplot[
            color= CBOrange,
            mark=*, 
            sharp plot, 
            line width=1.5pt
        ] coordinates {
            (1, 1.56) (2, 2.24) (4, 2.50)
        };
        \addlegendentry{RTA}
        
        \addplot[
            color=CBBlue,
            mark=none, 
            dashed,    
            line width=1.5pt
        ] coordinates {
            (0.5, 3.48) (4.5, 3.48) 
        };
        \addlegendentry{PGD}

        
        \end{axis}
    \end{tikzpicture}
         \caption{Processing time per image for RTA vs. PGD.}
    \label{fig:beam_search_time}
     \end{subfigure}
     \hfill
     \begin{subfigure}[b]{0.49\textwidth}
         \centering
         \begin{tikzpicture}
        \begin{axis}[
            width=\linewidth,
            height=4.5cm,
            xlabel={Step},
            ylabel={Target distance (km)},
            legend pos=north east,
            legend style={font=\scriptsize, cells={anchor=west}},
            ymajorgrids=true,
            xmajorgrids=true,
            grid style=dashed,
            tick label style={font=\normalsize},
            xmin=0, xmax=26,
            ymin=0, ymax=11000,
           tick label style={/pgf/number format/fixed},
        label style={font=\scriptsize},
        ticklabel style={font=\scriptsize},
        title style={font=\scriptsize},
        legend style={font=\scriptsize},
        ]
        
        \addplot[
            color=CBBlue,
        mark=*, 
        sharp plot, 
        line width=1.5pt
        ] coordinates {
            (0, 9735.35)
            (5, 5271.15)
            (10, 3037.18)
            (15, 2110.61)
            (20, 1708.16)
            (25, 1446.57)
        };
        \addlegendentry{PGD}
        
        \addplot[
            color=CBOrange,
        mark=square*, 
        sharp plot, 
        line width=1.5pt
        ] coordinates {
            (0, 9735.35)
            (5, 4469.92) 
            (10, 1344.77)
            (15, 516.48)
            (20, 186.52)
            (25, 55.00)
        };
        \addlegendentry{RTA}
        
        \end{axis}
    \end{tikzpicture}
         \caption{Target distance at different attack steps.}
         
         \label{fig:rta_pgd_comparison}
     \end{subfigure}
     \caption{Ablation study comparing RTA with PGD on IM2GPS3k, comparing average processing time (a) and target distance (b).}
     \label{fig:total_figure}
\end{figure}
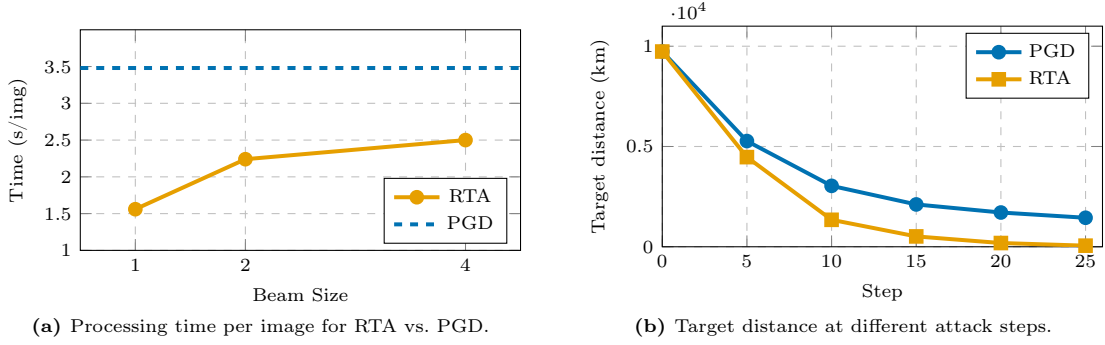

RTA requires more computation, but unlike PGD, most of the computation can be parallelized. The comparison in Fig. \ref{fig:beam_search_time} shows that RTA has a lower wall-clock time since the $J$ intermediate targets (see Alg. \ref{algo:roadtrip}) can be optimized in parallel. Recalling that in both cases we early-stop attacks when the target accuracy reaches 1km from the objective, this has a positive impact on our method, which, by searching a larger space, allows more early stops than PGD.\nn{We also report that GeoShield, in the same setting, runs at 20s/img, so 8x slower than RTA.} All RTA variants in Tab. \ref{tab:geoclip_im2gps} and Tab. \ref{tab:geoclip_yfcc} use beam size 4, \nn{5 parallel targets (respectively $K$ and $J$ in Alg.\ref{algo:roadtrip}), and for each of them 5 PGD steps. We use a step size of $\varepsilon/2$.
Additionally, we report that PGD uses around 4GB of VRAM, RTA around 6GB, and GeoShield around 8.5GB.}

\subsection{Ablation studies}
To better understand the behavior of RTA, we perform a series of ablation studies, focusing on the white-box setting.
We report a convergence analysis in Fig. \ref{fig:rta_pgd_comparison} against the strong PGD baseline. Here, we plot the distance from the target against the optimization step of the two methods. While PGD follows a more direct optimization path, RTA explores multiple paths, leading to a better solution.
As the attack progresses, RTA's beam search strategy identifies a more effective trajectory, ultimately converging to a solution significantly closer to the adversarial target than PGD. This confirms that the iterative, multi-path approach is more effective at navigating the feature space to find strong adversarial solutions.

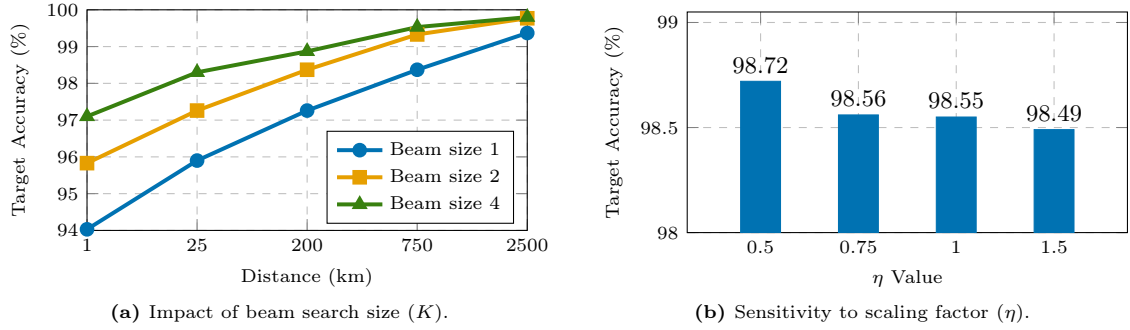
\begin{figure}[t]
    \centering
    
    \begin{subfigure}[b]{0.49\textwidth}
        \centering
        \begin{tikzpicture}
            \begin{axis}[
                width=\linewidth, 
                height=4.5cm,     
                xlabel={Distance (km)},
                ylabel={Target Accuracy (\%)},
                xmin=1, xmax=5,
                ymin=94, ymax=100,
                xtick={1, 2, 3, 4, 5},
                ytick={94, 95, 96, 97, 98, 99, 100},
                xticklabels={1, 25, 200, 750, 2500},
                legend pos=south east,
                ymajorgrids=true,
                xmajorgrids=true,
                grid style=dashed,
                tick label style={/pgf/number format/fixed},
                label style={font=\scriptsize},
                ticklabel style={font=\scriptsize},
                legend style={font=\scriptsize}
            ]
            \addplot[color=CBBlue, mark=*, line width=1.5pt] coordinates {(1, 94.03) (2, 95.90) (3, 97.26) (4, 98.37) (5, 99.37)};
            \addlegendentry{Beam size 1}
            
            \addplot[color=CBOrange, mark=square*, line width=1.5pt] coordinates {(1, 95.83) (2, 97.26) (3, 98.37) (4, 99.33) (5, 99.77)};
            \addlegendentry{Beam size 2}
            
            \addplot[color=CBGreen, mark=triangle*, line width=1.5pt] coordinates {(1, 97.10) (2, 98.30) (3, 98.87) (4, 99.53) (5, 99.80)};
            \addlegendentry{Beam size 4}
            \end{axis}
        \end{tikzpicture}
        \caption{Impact of beam search size ($K$).}
        \label{fig:beamsize}
    \end{subfigure}
    \hfill 
    \begin{subfigure}[b]{0.49\textwidth}
        \centering
        \begin{tikzpicture}
            \begin{axis}[
                width=\linewidth, 
                height=4.5cm,     
                ybar,
                enlarge x limits=0.25,
                ylabel={Target Accuracy (\%)},
                xlabel={$\eta$ Value},
                symbolic x coords={0.5, 0.75, 1, 1.5},
                xtick=data,
                tick align=inside,
                nodes near coords,
                nodes near coords align={vertical},
                node near coords style={font=\footnotesize},
                ymin=98, ymax=99.05, 
                grid=major,
                grid style=dashed,
                bar width=15pt,
                label style={font=\scriptsize},
                ticklabel style={font=\scriptsize}
            ]
            \addplot[fill=CBBlue, draw=CBBlue] coordinates {
                (0.5, 98.72) 
                (0.75, 98.56) 
                (1, 98.55) 
                (1.5, 98.49)
            };
            \end{axis}
        \end{tikzpicture}
        \caption{Sensitivity to scaling factor ($\eta$).}
        \label{fig:radius_mult}
    \end{subfigure}
    
    \caption{Experimental analysis of RTA parameters. (a) Larger beams increase target accuracy across distances. (b) Best average results are obtained with $\eta=0.5$.}
    \label{fig:combined_analysis}
\end{figure}

We then perform an ablation study on the beam size parameter ($K$ in Alg. \ref{algo:roadtrip}) of our proposed attack. We can observe from Fig.~\ref{fig:beamsize} a consistent trend showing that increasing the number of beams increases the attack's effectiveness. For the experiments reported throughout the paper we use K~=~4 beams.

In addition to the beam size, we conducted a further ablation study to analyze the sensitivity of the RoadTrip Attack to the sampling radius scaling factor, $\eta$. This hyperparameter governs the exploration scope of the intermediate target sampling at each step of the attack. As illustrated in Fig.~\ref{fig:radius_mult}, we evaluated the attack's performance with $\eta$ values set to 0.5, 0.75, 1.0, and 1.5. The empirical results indicate that a value of $\eta=0.5$ yields the highest target accuracy, particularly at the most challenging 1km and 25km thresholds. This suggests that a more conservative sampling strategy, which defines a tighter disk for intermediate targets relative to the remaining distance, allows the optimization to discover a more effective and precise adversarial path. Conversely, larger $\eta$ values, such as 1.5, result in a performance degradation. This behavior implies that excessive spatial jumps may hinder the optimization process, preventing it from converging to the best adversarial solution.

\vspace{-10pt}

\section{Conclusions}
In this work, we introduced the RoadTrip Attack, a novel adversarial method to address the privacy risks of image geolocalization. Our approach formulates the attack as an optimal distractor journey to a set of intermediate locations, using a beam search algorithm to find effective perturbation paths. Extensive experiments demonstrate that our method significantly outperforms both standard adversarial attacks and the only prior existing attack tailored to geolocalization. RTA achieves superior success rates \nn{in both white-box and black-box settings, particularly in low-budget regimes where perturbations must remain subtle.} 

The success of this journey-based approach highlights a critical vulnerability: the complex optimization landscape of geolocalization models can be more effectively navigated through a sequence of intermediate steps rather than a direct path.

\nn{Future work will address the robustness of our attack against preprocessing and purification-based countermeasures. In particular, we will consider naive JPEG compression, input-transformation defenses~\cite{xu2017feature}, reconstruction-based purification~\cite{song2018pixeldefend,meng2017magnet}, generative purification~\cite{samangouei2018defense,nie2022diffusion}, and recovery perturbations~\cite{jiang2023unlearnable}.}

\noindent\textbf{Acknowledgements.} This work was partially supported by the REG4AI project.

\bibliographystyle{splncs04}
\bibliography{main.bib}


\clearpage
\setcounter{page}{1}
\section*{Supplementary Material}
\setcounter{figure}{4}
\section{Adversarial Image Examples}
\label{sec:adv_examples}
\begin{figure*}[t]
    \centering
    \resizebox{.97\linewidth}{!}{
    \setlength{\tabcolsep}{4pt} 
        \begin{tabular}{c l l l l}
            & ~~~~\textbf{$\varepsilon=2/255$} & ~~~~\textbf{$\varepsilon=4/255$} & ~~~~\textbf{$\varepsilon=8/255$} & ~~~~\textbf{$\varepsilon=16/255$} \\
            \raisebox{0.6\height}{\rotatebox{90}{Geoshield\cite{geoshield}}}   & 
            \begin{tikzpicture}[spy using outlines={circle, red, line width=1pt, magnification=3, size=2.5cm, connect spies}]
                \node {\includegraphics[width=0.18\textwidth]{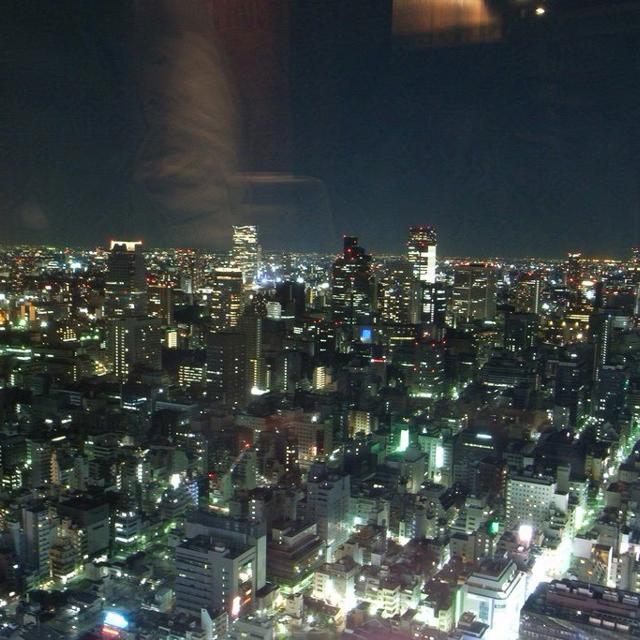}};
                \spy on (-0.4, 0.4) in node [right] at (0.1, -0.7);
            \end{tikzpicture} &
            \begin{tikzpicture}[spy using outlines={circle, red, line width=1pt, magnification=3, size=2.5cm, connect spies}]
                \node {\includegraphics[width=0.18\textwidth]{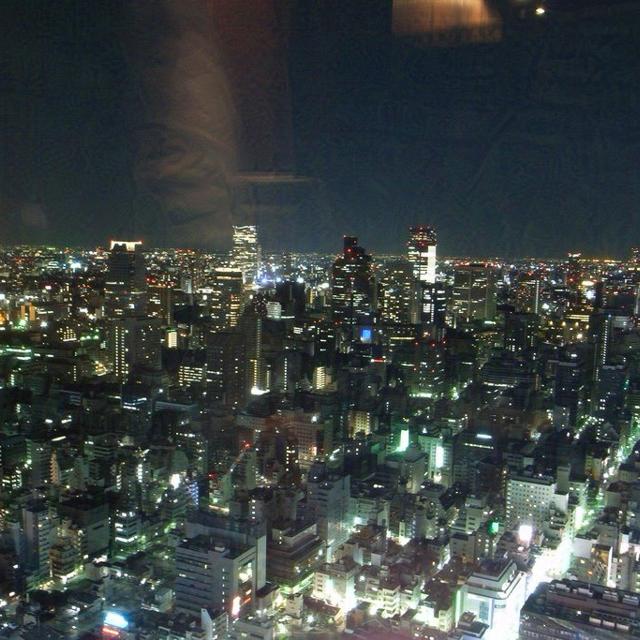}};
                \spy on (-0.4, 0.4) in node [right] at (0.1, -0.7);
            \end{tikzpicture} &
            \begin{tikzpicture}[spy using outlines={circle, red, line width=1pt, magnification=3, size=2.5cm, connect spies}]
                \node {\includegraphics[width=0.18\textwidth]{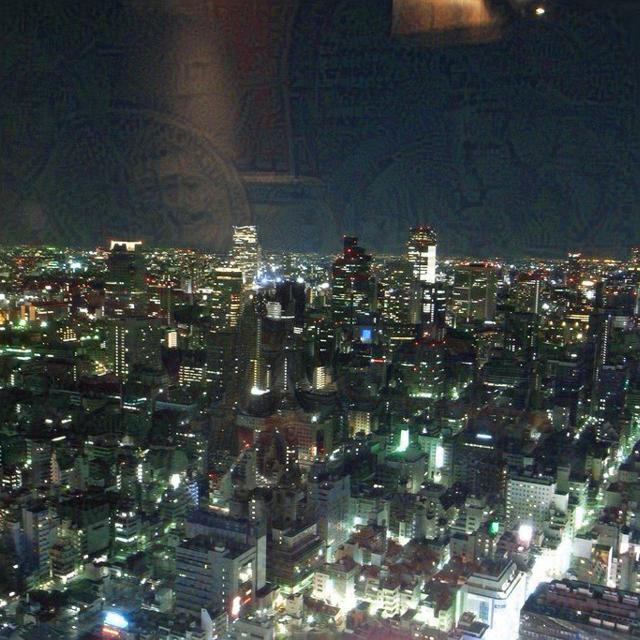}};
                \spy on (-0.4, 0.4) in node [right] at (0.1, -0.7);
            \end{tikzpicture} &
            \begin{tikzpicture}[spy using outlines={circle, red, line width=1pt, magnification=3, size=2.5cm, connect spies}]
                \node {\includegraphics[width=0.18\textwidth]{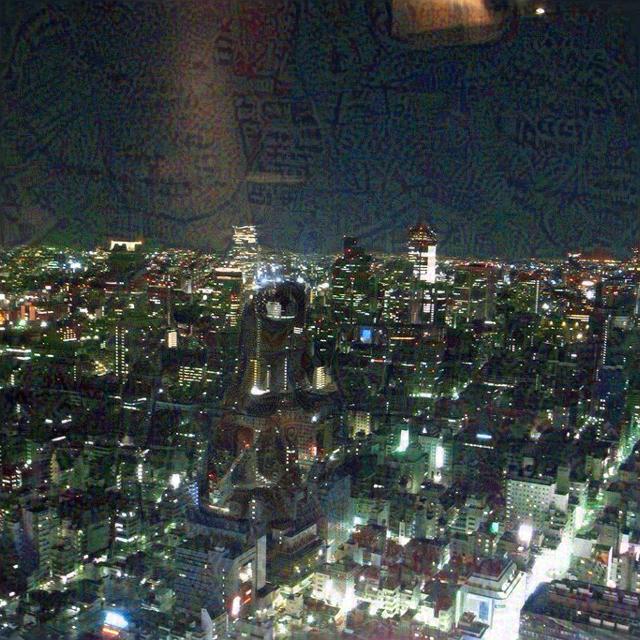}};
                \spy on (-0.4, 0.4) in node [right] at (0.1, -0.7);
            \end{tikzpicture} \\

          \raisebox{0.6\height}{\rotatebox{90}{~~~~~~~RTA}}   & 
            \begin{tikzpicture}[spy using outlines={circle, red, line width=1pt, magnification=3, size=2.5cm, connect spies}]
                \node {\includegraphics[width=0.18\textwidth]{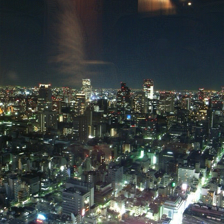}};
                \spy on (-0.4, 0.4) in node [right] at (0.1, -0.7);
            \end{tikzpicture} &
            \begin{tikzpicture}[spy using outlines={circle, red, line width=1pt, magnification=3, size=2.5cm, connect spies}]
                \node {\includegraphics[width=0.18\textwidth]{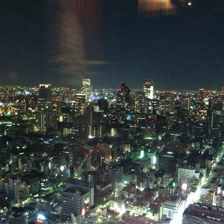}};
                \spy on (-0.4, 0.4) in node [right] at (0.1, -0.7);
            \end{tikzpicture} &
            \begin{tikzpicture}[spy using outlines={circle, red, line width=1pt, magnification=3, size=2.5cm, connect spies}]
                \node {\includegraphics[width=0.18\textwidth]{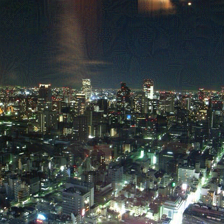}};
                \spy on (-0.4, 0.4) in node [right] at (0.1, -0.7);
            \end{tikzpicture} &
            \begin{tikzpicture}[spy using outlines={circle, red, line width=1pt, magnification=3, size=2.5cm, connect spies}]
                \node {\includegraphics[width=0.18\textwidth]{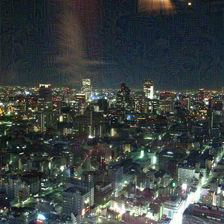}};
                \spy on (-0.4, 0.4) in node [right] at (0.1, -0.7);
            \end{tikzpicture} \\    

            & ~~~~\textbf{$\varepsilon=2/255$} & ~~~~\textbf{$\varepsilon=4/255$} & ~~~~\textbf{$\varepsilon=8/255$} & ~~~~\textbf{$\varepsilon=16/255$} \\


            \raisebox{0.6\height}{\rotatebox{90}{Geoshield\cite{geoshield}}}   & 
            \begin{tikzpicture}[spy using outlines={circle, red, line width=1pt, magnification=3, size=2.5cm, connect spies}]
                \node {\includegraphics[width=0.18\textwidth]{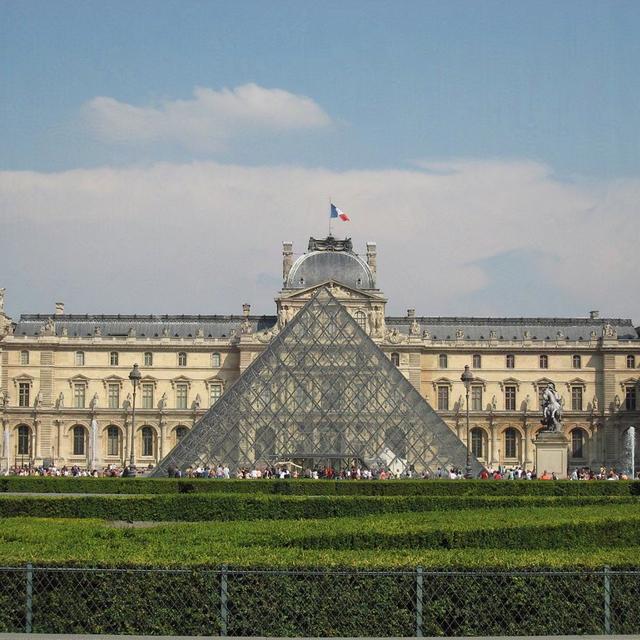}};
                \spy on (-0.4, 0.6) in node [right] at (0.1, -0.7);
            \end{tikzpicture} &
            \begin{tikzpicture}[spy using outlines={circle, red, line width=1pt, magnification=3, size=2.5cm, connect spies}]
                \node {\includegraphics[width=0.18\textwidth]{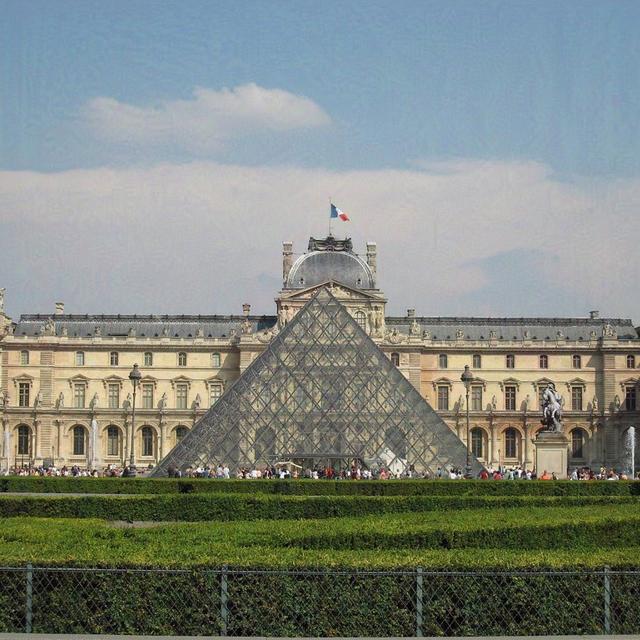}};
                \spy on (-0.4, 0.6) in node [right] at (0.1, -0.7);
            \end{tikzpicture} &
            \begin{tikzpicture}[spy using outlines={circle, red, line width=1pt, magnification=3, size=2.5cm, connect spies}]
                \node {\includegraphics[width=0.18\textwidth]{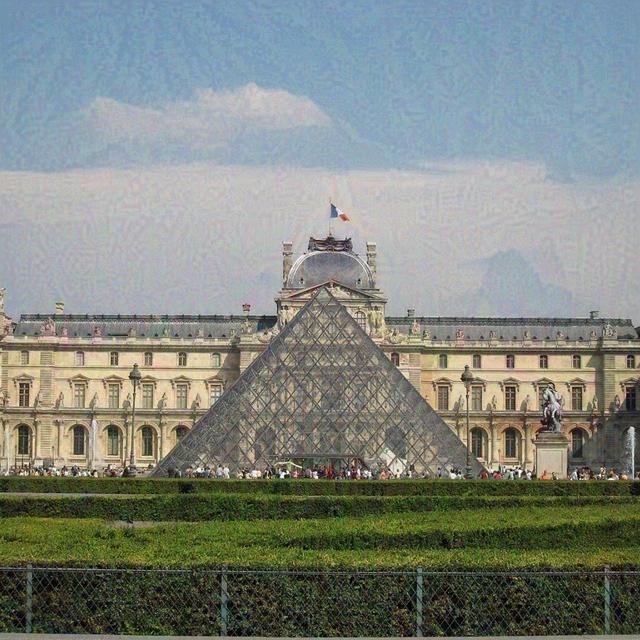}};
                \spy on (-0.4, 0.6) in node [right] at (0.1, -0.7);
            \end{tikzpicture} &
            \begin{tikzpicture}[spy using outlines={circle, red, line width=1pt, magnification=3, size=2.5cm, connect spies}]
                \node {\includegraphics[width=0.18\textwidth]{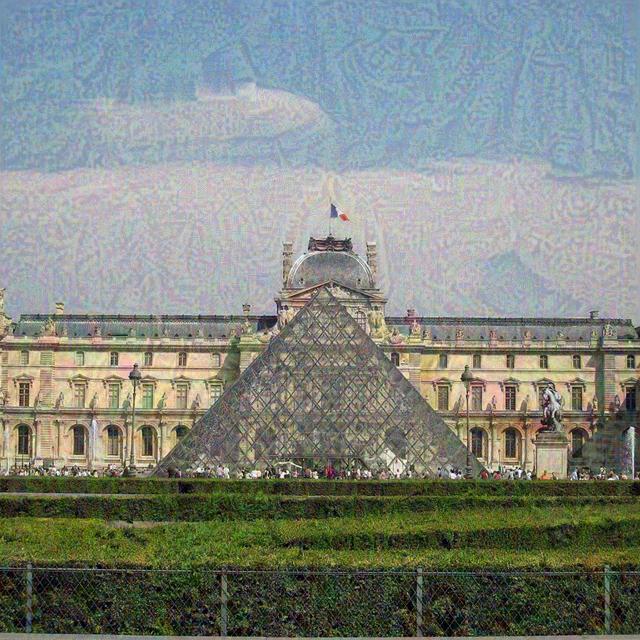}};
                \spy on (-0.4, 0.6) in node [right] at (0.1, -0.7);
            \end{tikzpicture} \\

          \raisebox{0.6\height}{\rotatebox{90}{~~~~~~~RTA}}   & 
            \begin{tikzpicture}[spy using outlines={circle, red, line width=1pt, magnification=3, size=2.5cm, connect spies}]
                \node {\includegraphics[width=0.18\textwidth]{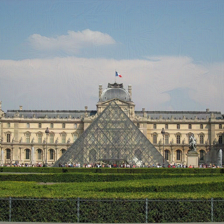}};
                \spy on (-0.4, 0.6) in node [right] at (0.1, -0.7);
            \end{tikzpicture} &
            \begin{tikzpicture}[spy using outlines={circle, red, line width=1pt, magnification=3, size=2.5cm, connect spies}]
                \node {\includegraphics[width=0.18\textwidth]{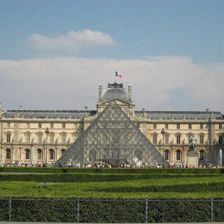}};
                \spy on (-0.4, 0.6) in node [right] at (0.1, -0.7);
            \end{tikzpicture} &
            \begin{tikzpicture}[spy using outlines={circle, red, line width=1pt, magnification=3, size=2.5cm, connect spies}]
                \node {\includegraphics[width=0.18\textwidth]{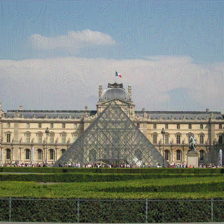}};
                \spy on (-0.4, 0.6) in node [right] at (0.1, -0.7);
            \end{tikzpicture} &
            \begin{tikzpicture}[spy using outlines={circle, red, line width=1pt, magnification=3, size=2.5cm, connect spies}]
                \node {\includegraphics[width=0.18\textwidth]{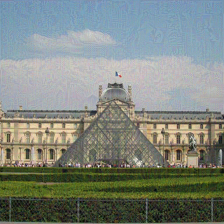}};
                \spy on (-0.4, 0.6) in node [right] at (0.1, -0.7);
            \end{tikzpicture} \\ 

        & ~~~~\textbf{$\varepsilon=2/255$} & ~~~~\textbf{$\varepsilon=4/255$} & ~~~~\textbf{$\varepsilon=8/255$} & ~~~~\textbf{$\varepsilon=16/255$} \\


            \raisebox{0.6\height}{\rotatebox{90}{Geoshield\cite{geoshield}}}   & 
            \begin{tikzpicture}[spy using outlines={circle, red, line width=1pt, magnification=3, size=2.5cm, connect spies}]
                \node {\includegraphics[width=0.18\textwidth]{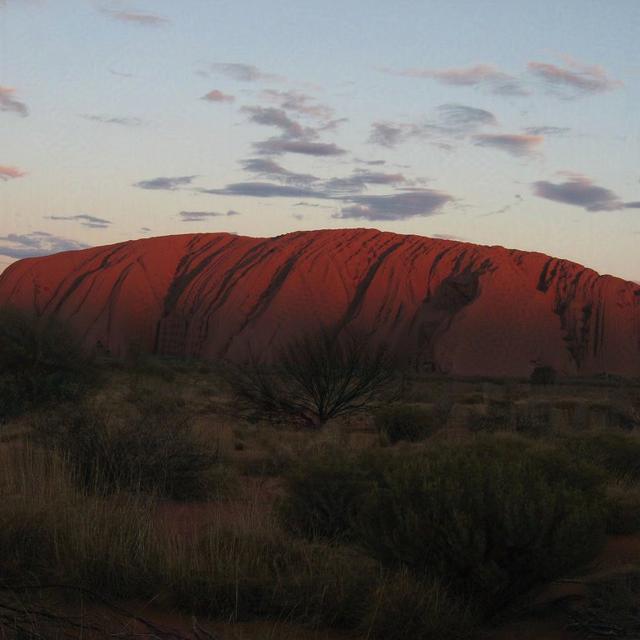}};
                \spy on (-0.4, 0.2) in node [right] at (0.1, -0.7);
            \end{tikzpicture} &
            \begin{tikzpicture}[spy using outlines={circle, red, line width=1pt, magnification=3, size=2.5cm, connect spies}]
                \node {\includegraphics[width=0.18\textwidth]{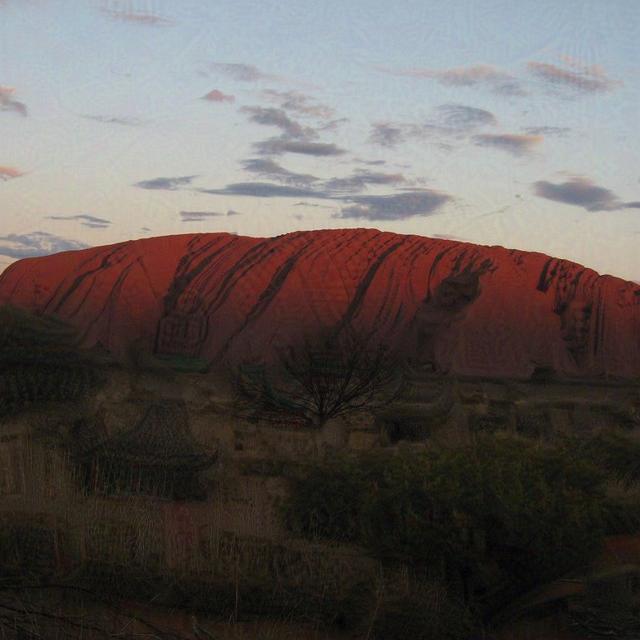}};
                \spy on (-0.4, 0.2) in node [right] at (0.1, -0.7);
            \end{tikzpicture} &
            \begin{tikzpicture}[spy using outlines={circle, red, line width=1pt, magnification=3, size=2.5cm, connect spies}]
                \node {\includegraphics[width=0.18\textwidth]{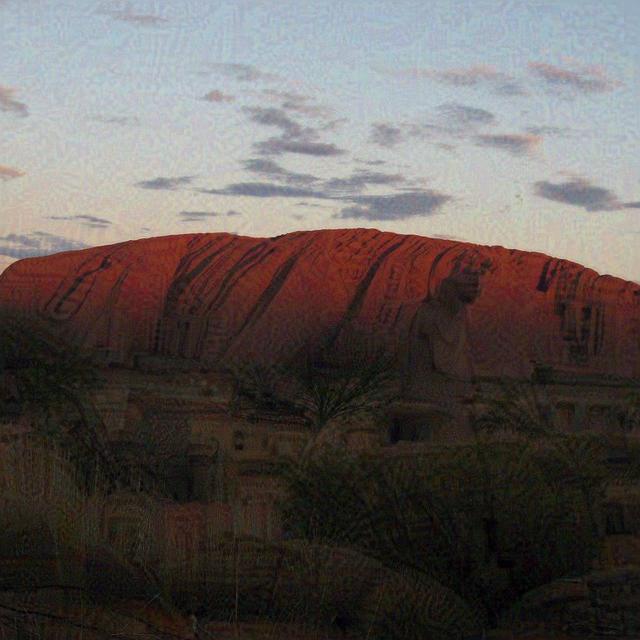}};
                \spy on (-0.4, 0.2) in node [right] at (0.1, -0.7);
            \end{tikzpicture} &
            \begin{tikzpicture}[spy using outlines={circle, red, line width=1pt, magnification=3, size=2.5cm, connect spies}]
                \node {\includegraphics[width=0.18\textwidth]{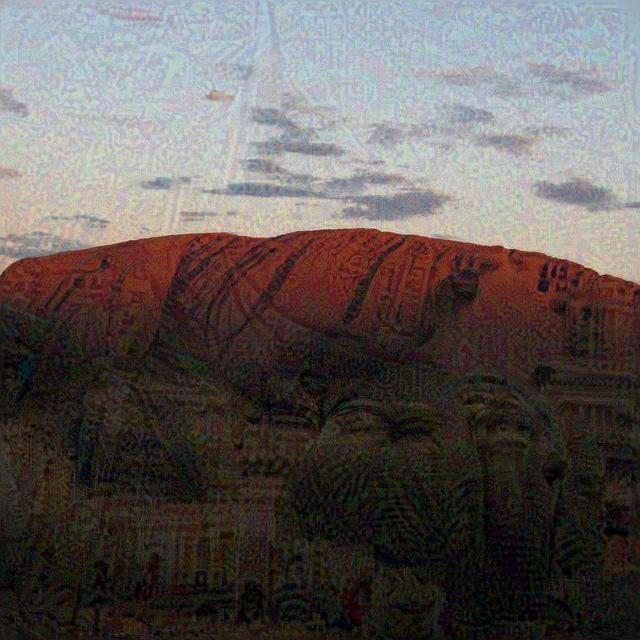}};
                \spy on (-0.4, 0.2) in node [right] at (0.1, -0.7);
            \end{tikzpicture} \\

          \raisebox{0.6\height}{\rotatebox{90}{~~~~~~~RTA}}   & 
            \begin{tikzpicture}[spy using outlines={circle, red, line width=1pt, magnification=3, size=2.5cm, connect spies}]
                \node {\includegraphics[width=0.18\textwidth]{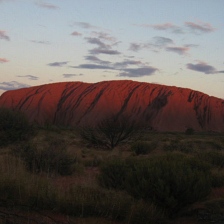}};
                \spy on (-0.4, 0.2) in node [right] at (0.1, -0.7);
            \end{tikzpicture} &
            \begin{tikzpicture}[spy using outlines={circle, red, line width=1pt, magnification=3, size=2.5cm, connect spies}]
                \node {\includegraphics[width=0.18\textwidth]{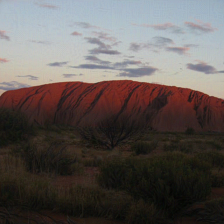}};
                \spy on (-0.4, 0.2) in node [right] at (0.1, -0.7);
            \end{tikzpicture} &
            \begin{tikzpicture}[spy using outlines={circle, red, line width=1pt, magnification=3, size=2.5cm, connect spies}]
                \node {\includegraphics[width=0.18\textwidth]{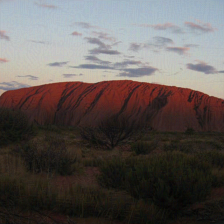}};
                \spy on (-0.4, 0.2) in node [right] at (0.1, -0.7);
            \end{tikzpicture} &
            \begin{tikzpicture}[spy using outlines={circle, red, line width=1pt, magnification=3, size=2.5cm, connect spies}]
                \node {\includegraphics[width=0.18\textwidth]{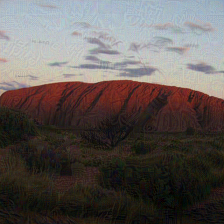}};
                \spy on (-0.4, 0.2) in node [right] at (0.1, -0.7);
            \end{tikzpicture} \\

        \end{tabular}
        }
    \caption{Visual comparison of adversarial noise for four budgets $\varepsilon = 2/255$, $4/255$, $8/255$, $16/255$ and two methods (Geoshield and ours). The magnified area highlights the fine-grained perturbation patterns for the attack. Best viewed in color on a screen.}
    \label{fig:SUPPL_comparison}
\end{figure*}

In Fig.~\ref{fig:ours_geoshield_comparison} of the main paper, we showed a single adversarial example due to space limitations; in Fig.~\ref{fig:SUPPL_comparison}, we show additional attacked images with zoomed-in crops to highlight the difference in noise patterns between our method and the competing method GeoShield\cite{geoshield}.

\section{Internal Representation Analysis}
\begin{figure}[h]
    \centering
    \resizebox{\columnwidth}{!}{
        \begin{tikzpicture}[
            box/.style={rectangle, draw, thick, minimum size=2.5cm, inner sep=0pt},
            label style/.style={rotate=90, font=\large, anchor=south}, 
            header style/.style={font=\sffamily\Large\bfseries, anchor=south}
        ]

            \node[header style] at (0, 2.0) {Query};
            \node[header style] at (9, 2.0) {Top-5 Similar Images};

            \node[label style] at (-1.8, 0) {Ground Truth};
            \node[box] at (0,0) {\includegraphics[width=2.5cm, height=2.5cm]{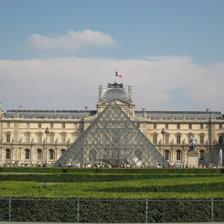}}; 
            
            \node[box] at (3, 0) {\includegraphics[width=2.5cm, height=2.5cm]{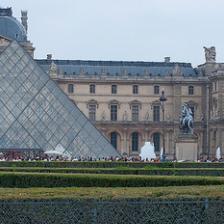}};
            \node[box] at (6, 0) {\includegraphics[width=2.5cm, height=2.5cm]{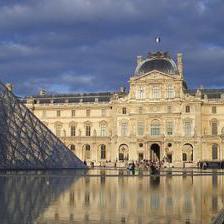}};
            \node[box] at (9, 0) {\includegraphics[width=2.5cm, height=2.5cm]{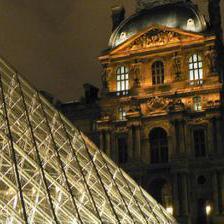}};
            \node[box] at (12, 0) {\includegraphics[width=2.5cm, height=2.5cm]{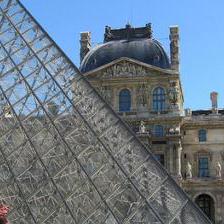}};
            \node[box] at (15, 0) {\includegraphics[width=2.5cm, height=2.5cm]{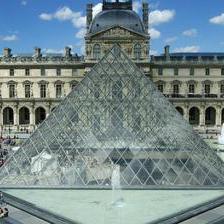}};
            \node[label style] at (-1.8, -3.2) {GeoShield\cite{geoshield}};
            \node[box] at (0,-3.2) {\includegraphics[width=2.5cm, height=2.5cm]{suppl_eccv/reference/geoshield/eps2/524999707_5413fe3302_246_80246871_N00.jpg}};
            
            \node[box] at (3, -3.2) {\includegraphics[width=2.5cm, height=2.5cm]{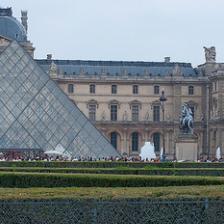}};
            \node[box] at (6, -3.2) {\includegraphics[width=2.5cm, height=2.5cm]{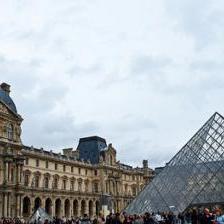}};
            \node[box] at (9, -3.2) {\includegraphics[width=2.5cm, height=2.5cm]{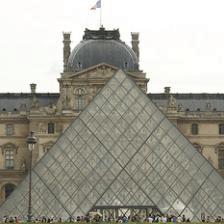}};
            \node[box] at (12, -3.2) {\includegraphics[width=2.5cm, height=2.5cm]{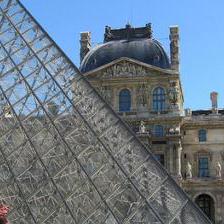}};
            \node[box] at (15, -3.2) {\includegraphics[width=2.5cm, height=2.5cm]{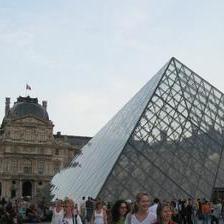}};
            \node[label style] at (-1.8, -6.4) {RTA};
            \node[box] at (0,-6.4) {\includegraphics[width=2.5cm, height=2.5cm]{suppl_eccv/reference/rta/eps2/524999707_5413fe3302_246_80246871_N00.png}};
            
            \node[box] at (3, -6.4) {\includegraphics[width=2.5cm, height=2.5cm]{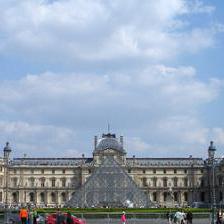}};
            \node[box] at (6, -6.4) {\includegraphics[width=2.5cm, height=2.5cm]{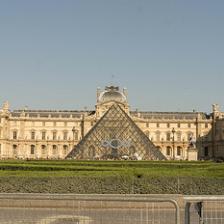}};
            \node[box] at (9, -6.4) {\includegraphics[width=2.5cm, height=2.5cm]{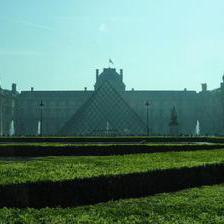}};
            \node[box] at (12, -6.4) {\includegraphics[width=2.5cm, height=2.5cm]{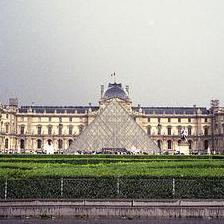}};
            \node[box] at (15, -6.4) {\includegraphics[width=2.5cm, height=2.5cm]{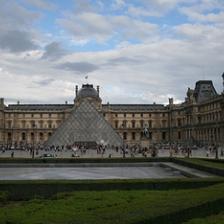}};
            \draw[thick, dotted] (1.5, 2.8) -- (1.5, -8.0);

        \end{tikzpicture}
    }
    \caption{Example of retrieved images by Img2Loc\cite{img2loc}. $\varepsilon = 2/255$}
    \label{fig:img2loc_eps2_mountain}
\end{figure}

\begin{figure}[h]
    \centering
    \resizebox{\columnwidth}{!}{
        \begin{tikzpicture}[
            box/.style={rectangle, draw, thick, minimum size=2.5cm, inner sep=0pt},
            label style/.style={rotate=90, font=\large, anchor=south}, 
            header style/.style={font=\sffamily\Large\bfseries, anchor=south}
        ]

            \node[header style] at (0, 2.0) {Query};
            \node[header style] at (9, 2.0) {Top-5 Similar Images};

            \node[label style] at (-1.8, 0) {Ground Truth};
            \node[box] at (0,0) {\includegraphics[width=2.5cm, height=2.5cm]{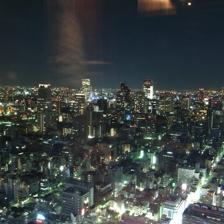}}; 
            
            \node[box] at (3, 0) {\includegraphics[width=2.5cm, height=2.5cm]{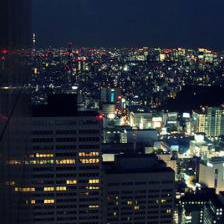}};
            \node[box] at (6, 0) {\includegraphics[width=2.5cm, height=2.5cm]{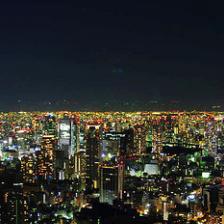}};
            \node[box] at (9, 0) {\includegraphics[width=2.5cm, height=2.5cm]{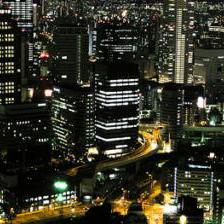}};
            \node[box] at (12, 0) {\includegraphics[width=2.5cm, height=2.5cm]{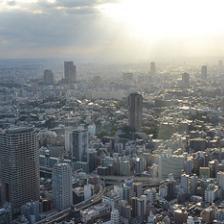}};
            \node[box] at (15, 0) {\includegraphics[width=2.5cm, height=2.5cm]{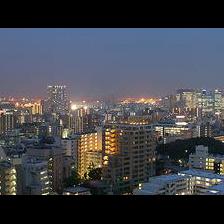}};
            \node[label style] at (-1.8, -3.2) {GeoShield\cite{geoshield}};
            \node[box] at (0,-3.2) {\includegraphics[width=2.5cm, height=2.5cm]{suppl_eccv/reference/geoshield/eps4/462922891_67372e6aed_196_25159586_N00.jpg}};
            
            \node[box] at (3, -3.2) {\includegraphics[width=2.5cm, height=2.5cm]{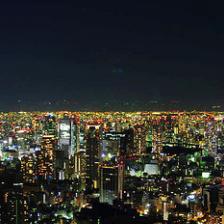}};
            \node[box] at (6, -3.2) {\includegraphics[width=2.5cm, height=2.5cm]{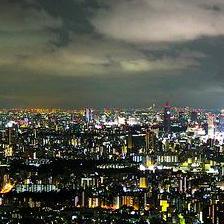}};
            \node[box] at (9, -3.2) {\includegraphics[width=2.5cm, height=2.5cm]{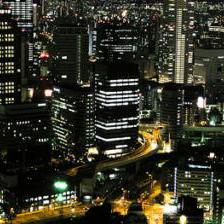}};
            \node[box] at (12, -3.2) {\includegraphics[width=2.5cm, height=2.5cm]{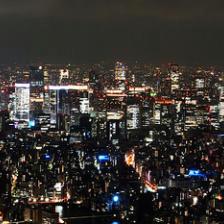}};
            \node[box] at (15, -3.2) {\includegraphics[width=2.5cm, height=2.5cm]{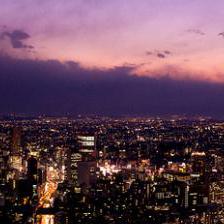}};
            \node[label style] at (-1.8, -6.4) {RTA};
            \node[box] at (0,-6.4) {\includegraphics[width=2.5cm, height=2.5cm]{suppl_eccv/reference/rta/eps4/462922891_67372e6aed_196_25159586_N00.png}};
            
            \node[box] at (3, -6.4) {\includegraphics[width=2.5cm, height=2.5cm]{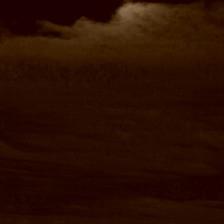}};
            \node[box] at (6, -6.4) {\includegraphics[width=2.5cm, height=2.5cm]{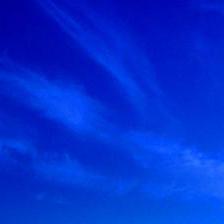}};
            \node[box] at (9, -6.4) {\includegraphics[width=2.5cm, height=2.5cm]{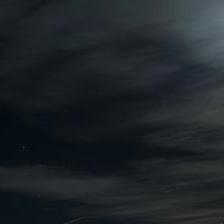}};
            \node[box] at (12, -6.4) {\includegraphics[width=2.5cm, height=2.5cm]{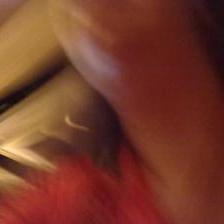}};
            \node[box] at (15, -6.4) {\includegraphics[width=2.5cm, height=2.5cm]{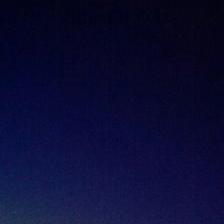}};
            \draw[thick, dotted] (1.5, 2.8) -- (1.5, -8.0);

        \end{tikzpicture}
    }
    \caption{Example of retrieved images by Img2Loc\cite{img2loc}. $\varepsilon = 4/255$}
    \label{fig:img2loc_eps4_tokyo}
\end{figure}

\begin{figure}[h]
    \centering
    \resizebox{\columnwidth}{!}{
        \begin{tikzpicture}[
            box/.style={rectangle, draw, thick, minimum size=2.5cm, inner sep=0pt},
            label style/.style={rotate=90, font=\large, anchor=south}, 
            header style/.style={font=\sffamily\Large\bfseries, anchor=south}
        ]

            \node[header style] at (0, 2.0) {Query};
            \node[header style] at (9, 2.0) {Top-5 Similar Images};

            \node[label style] at (-1.8, 0) {Ground Truth};
            \node[box] at (0,0) {\includegraphics[width=2.5cm, height=2.5cm]{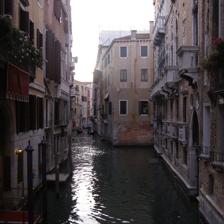}}; 
            
            \node[box] at (3, 0) {\includegraphics[width=2.5cm, height=2.5cm]{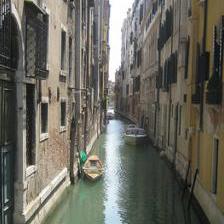}};
            \node[box] at (6, 0) {\includegraphics[width=2.5cm, height=2.5cm]{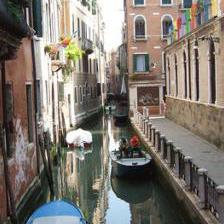}};
            \node[box] at (9, 0) {\includegraphics[width=2.5cm, height=2.5cm]{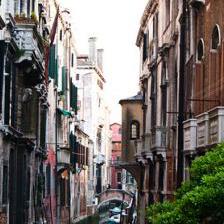}};
            \node[box] at (12, 0) {\includegraphics[width=2.5cm, height=2.5cm]{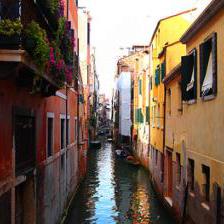}};
            \node[box] at (15, 0) {\includegraphics[width=2.5cm, height=2.5cm]{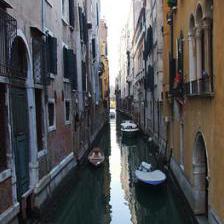}};
            \node[label style] at (-1.8, -3.2) {GeoShield\cite{geoshield}};
            \node[box] at (0,-3.2) {\includegraphics[width=2.5cm, height=2.5cm]{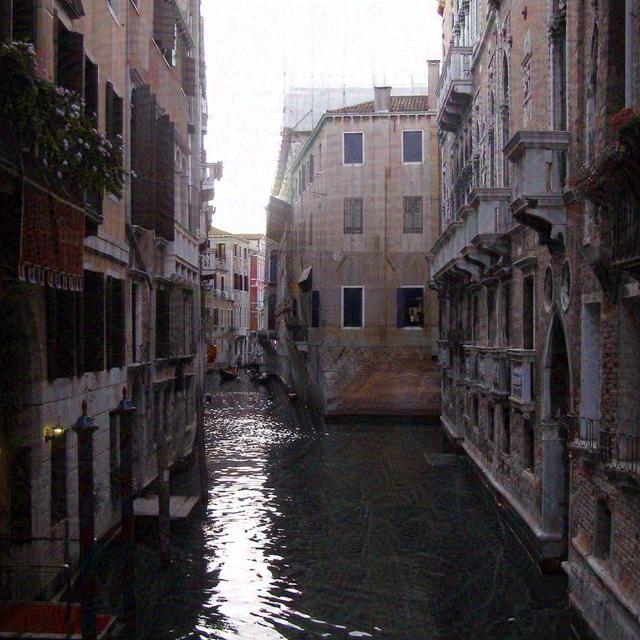}};
            
            \node[box] at (3, -3.2) {\includegraphics[width=2.5cm, height=2.5cm]{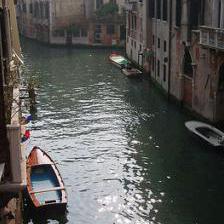}};
            \node[box] at (6, -3.2) {\includegraphics[width=2.5cm, height=2.5cm]{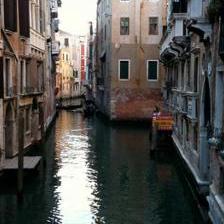}};
            \node[box] at (9, -3.2) {\includegraphics[width=2.5cm, height=2.5cm]{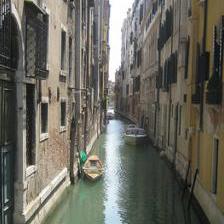}};
            \node[box] at (12, -3.2) {\includegraphics[width=2.5cm, height=2.5cm]{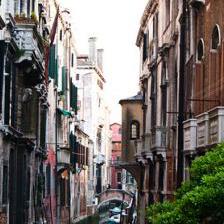}};
            \node[box] at (15, -3.2) {\includegraphics[width=2.5cm, height=2.5cm]{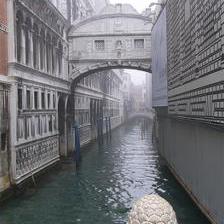}};
            \node[label style] at (-1.8, -6.4) {RTA};
            \node[box] at (0,-6.4) {\includegraphics[width=2.5cm, height=2.5cm]{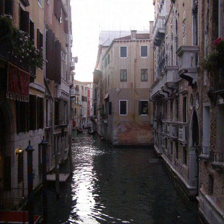}};
            
            \node[box] at (3, -6.4) {\includegraphics[width=2.5cm, height=2.5cm]{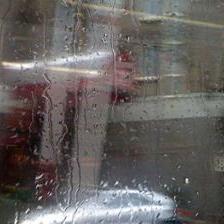}};
            \node[box] at (6, -6.4) {\includegraphics[width=2.5cm, height=2.5cm]{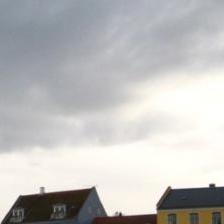}};
            \node[box] at (9, -6.4) {\includegraphics[width=2.5cm, height=2.5cm]{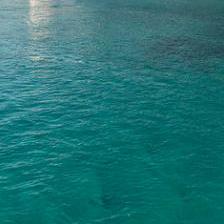}};
            \node[box] at (12, -6.4) {\includegraphics[width=2.5cm, height=2.5cm]{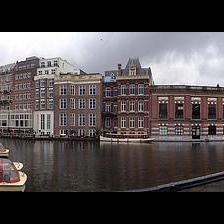}};
            \node[box] at (15, -6.4) {\includegraphics[width=2.5cm, height=2.5cm]{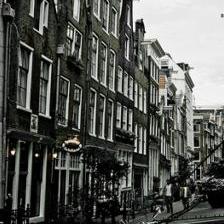}};
            \draw[thick, dotted] (1.5, 2.8) -- (1.5, -8.0);

        \end{tikzpicture}
    }
    \caption{Example of retrieved images by Img2Loc\cite{img2loc}. $\varepsilon = 8/255$}
    \label{fig:img2loc_eps8_venice}
\end{figure}

\begin{figure}[h]
    \centering
    \resizebox{\columnwidth}{!}{
        \begin{tikzpicture}[
            box/.style={rectangle, draw, thick, minimum size=2.5cm, inner sep=0pt},
            label style/.style={rotate=90, font=\large, anchor=south}, 
            header style/.style={font=\sffamily\Large\bfseries, anchor=south}
        ]

            \node[header style] at (0, 2.0) {Query};
            \node[header style] at (9, 2.0) {Top-5 Similar Images};

            \node[label style] at (-1.8, 0) {Ground Truth};
            \node[box] at (0,0) {\includegraphics[width=2.5cm, height=2.5cm]{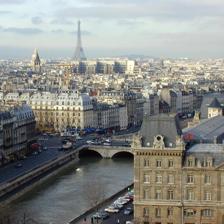}}; 
            
            \node[box] at (3, 0) {\includegraphics[width=2.5cm, height=2.5cm]{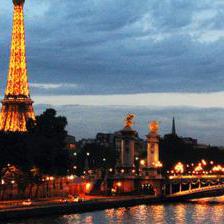}};
            \node[box] at (6, 0) {\includegraphics[width=2.5cm, height=2.5cm]{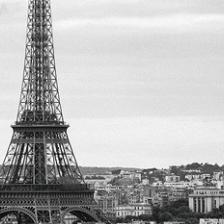}};
            \node[box] at (9, 0) {\includegraphics[width=2.5cm, height=2.5cm]{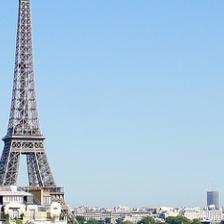}};
            \node[box] at (12, 0) {\includegraphics[width=2.5cm, height=2.5cm]{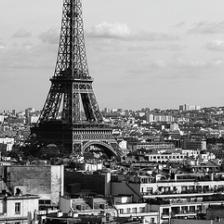}};
            \node[box] at (15, 0) {\includegraphics[width=2.5cm, height=2.5cm]{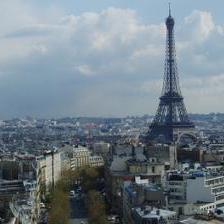}};
            \node[label style] at (-1.8, -3.2) {GeoShield\cite{geoshield}};
            \node[box] at (0,-3.2) {\includegraphics[width=2.5cm, height=2.5cm]{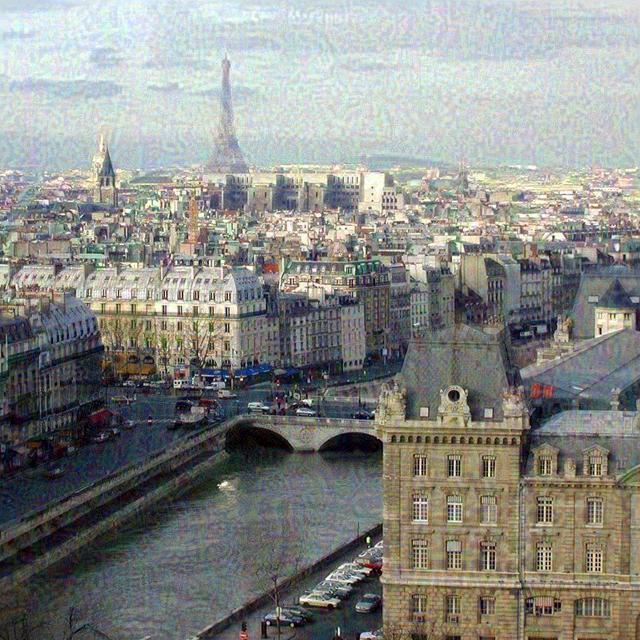}};
            
            \node[box] at (3, -3.2) {\includegraphics[width=2.5cm, height=2.5cm]{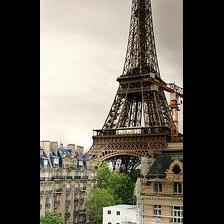}};
            \node[box] at (6, -3.2) {\includegraphics[width=2.5cm, height=2.5cm]{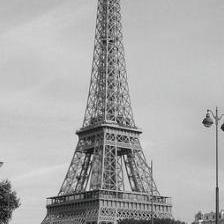}};
            \node[box] at (9, -3.2) {\includegraphics[width=2.5cm, height=2.5cm]{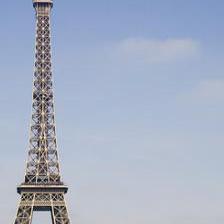}};
            \node[box] at (12, -3.2) {\includegraphics[width=2.5cm, height=2.5cm]{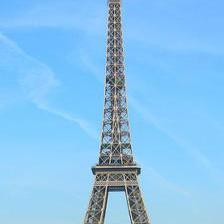}};
            \node[box] at (15, -3.2) {\includegraphics[width=2.5cm, height=2.5cm]{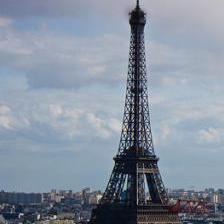}};
            \node[label style] at (-1.8, -6.4) {RTA};
            \node[box] at (0,-6.4) {\includegraphics[width=2.5cm, height=2.5cm]{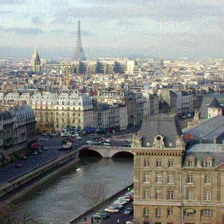}};
            
            \node[box] at (3, -6.4) {\includegraphics[width=2.5cm, height=2.5cm]{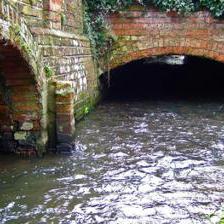}};
            \node[box] at (6, -6.4) {\includegraphics[width=2.5cm, height=2.5cm]{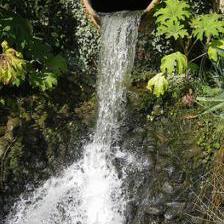}};
            \node[box] at (9, -6.4) {\includegraphics[width=2.5cm, height=2.5cm]{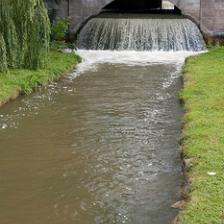}};
            \node[box] at (12, -6.4) {\includegraphics[width=2.5cm, height=2.5cm]{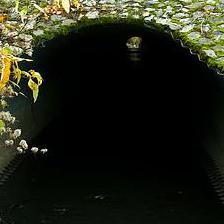}};
            \node[box] at (15, -6.4) {\includegraphics[width=2.5cm, height=2.5cm]{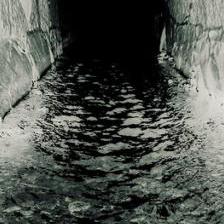}};
            \draw[thick, dotted] (1.5, 2.8) -- (1.5, -8.0);

        \end{tikzpicture}
    }
    \caption{Example of retrieved images by Img2Loc\cite{img2loc}. $\varepsilon = 16/255$}
    \label{fig:img2loc_eps16_parigi}
\end{figure}

\begin{figure}[h]
    \centering
    \resizebox{\columnwidth}{!}{
        \begin{tikzpicture}[
            box/.style={rectangle, draw, thick, minimum size=2.5cm, inner sep=0pt},
            label style/.style={rotate=90, font=\large, anchor=south}, 
            header style/.style={font=\sffamily\Large\bfseries, anchor=south}
        ]

            \node[header style] at (0, 2.0) {Query};
            \node[header style] at (9, 2.0) {Top-5 Similar Images};

            \node[label style] at (-1.8, 0) {Ground Truth};
            \node[box] at (0,0) {\includegraphics[width=2.5cm, height=2.5cm]{suppl_eccv/reference/clean/524999707_5413fe3302_246_80246871_N00.jpg}}; 
            
            \node[box] at (3, 0) {\includegraphics[width=2.5cm, height=2.5cm]{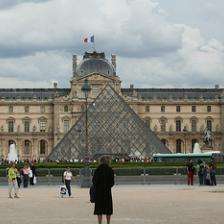}};
            \node[box] at (6, 0) {\includegraphics[width=2.5cm, height=2.5cm]{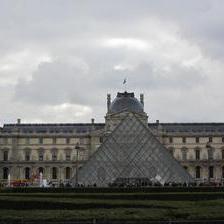}};
            \node[box] at (9, 0) {\includegraphics[width=2.5cm, height=2.5cm]{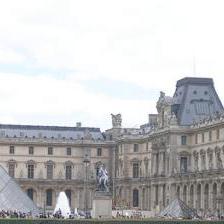}};
            \node[box] at (12, 0) {\includegraphics[width=2.5cm, height=2.5cm]{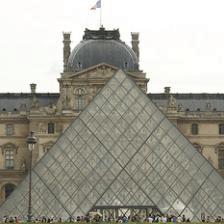}};
            \node[box] at (15, 0) {\includegraphics[width=2.5cm, height=2.5cm]{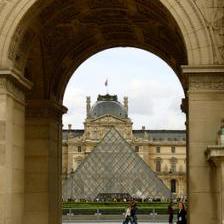}};
            \node[label style] at (-1.8, -3.2) {GeoShield\cite{geoshield}};
            \node[box] at (0,-3.2) {\includegraphics[width=2.5cm, height=2.5cm]{suppl_eccv/reference/geoshield/eps2/524999707_5413fe3302_246_80246871_N00.jpg}};
            
            \node[box] at (3, -3.2) {\includegraphics[width=2.5cm, height=2.5cm]{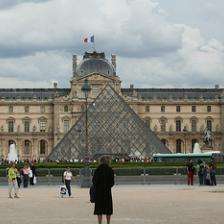}};
            \node[box] at (6, -3.2) {\includegraphics[width=2.5cm, height=2.5cm]{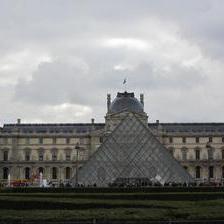}};
            \node[box] at (9, -3.2) {\includegraphics[width=2.5cm, height=2.5cm]{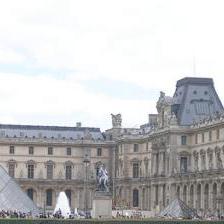}};
            \node[box] at (12, -3.2) {\includegraphics[width=2.5cm, height=2.5cm]{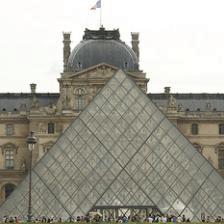}};
            \node[box] at (15, -3.2) {\includegraphics[width=2.5cm, height=2.5cm]{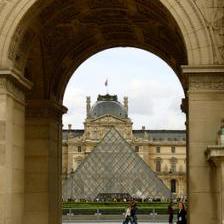}};
            \node[label style] at (-1.8, -6.4) {RTA};
            \node[box] at (0,-6.4) {\includegraphics[width=2.5cm, height=2.5cm]{suppl_eccv/reference/rta/eps2/524999707_5413fe3302_246_80246871_N00.png}};
            
            \node[box] at (3, -6.4) {\includegraphics[width=2.5cm, height=2.5cm]{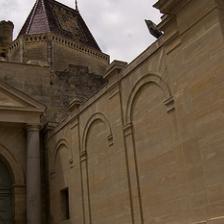}};
            \node[box] at (6, -6.4) {\includegraphics[width=2.5cm, height=2.5cm]{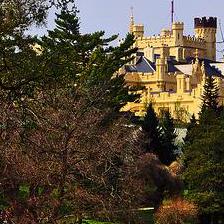}};
            \node[box] at (9, -6.4) {\includegraphics[width=2.5cm, height=2.5cm]{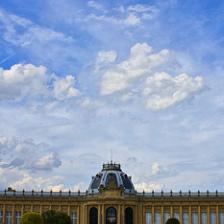}};
            \node[box] at (12, -6.4) {\includegraphics[width=2.5cm, height=2.5cm]{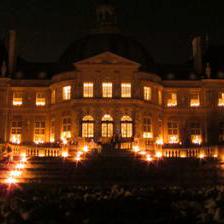}};
            \node[box] at (15, -6.4) {\includegraphics[width=2.5cm, height=2.5cm]{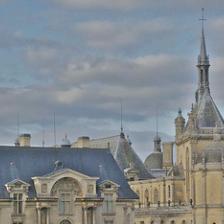}};
            \draw[thick, dotted] (1.5, 2.8) -- (1.5, -8.0);

        \end{tikzpicture}
    }
    \caption{Example of retrieved images by G3\cite{jia2024g3}. $\varepsilon = 2/255$}
    \label{fig:g3_eps2_mountain}
\end{figure}

\begin{figure}[h]
    \centering
    \resizebox{\columnwidth}{!}{
        \begin{tikzpicture}[
            box/.style={rectangle, draw, thick, minimum size=2.5cm, inner sep=0pt},
            label style/.style={rotate=90, font=\large, anchor=south}, 
            header style/.style={font=\sffamily\Large\bfseries, anchor=south}
        ]

            \node[header style] at (0, 2.0) {Query};
            \node[header style] at (9, 2.0) {Top-5 Similar Images};

            \node[label style] at (-1.8, 0) {Ground Truth};
            \node[box] at (0,0) {\includegraphics[width=2.5cm, height=2.5cm]{suppl_eccv/reference/clean/462922891_67372e6aed_196_25159586_N00.jpg}}; 
            
            \node[box] at (3, 0) {\includegraphics[width=2.5cm, height=2.5cm]{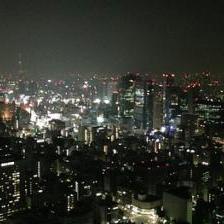}};
            \node[box] at (6, 0) {\includegraphics[width=2.5cm, height=2.5cm]{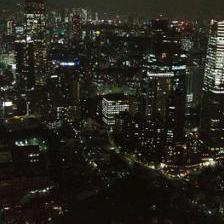}};
            \node[box] at (9, 0) {\includegraphics[width=2.5cm, height=2.5cm]{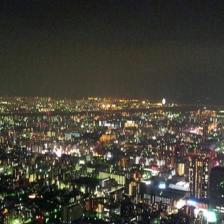}};
            \node[box] at (12, 0) {\includegraphics[width=2.5cm, height=2.5cm]{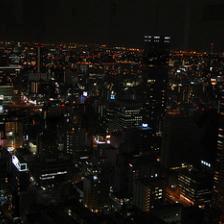}};
            \node[box] at (15, 0) {\includegraphics[width=2.5cm, height=2.5cm]{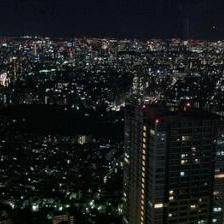}};
            \node[label style] at (-1.8, -3.2) {GeoShield\cite{geoshield}};
            \node[box] at (0,-3.2) {\includegraphics[width=2.5cm, height=2.5cm]{suppl_eccv/reference/geoshield/eps4/462922891_67372e6aed_196_25159586_N00.jpg}};
            
            \node[box] at (3, -3.2) {\includegraphics[width=2.5cm, height=2.5cm]{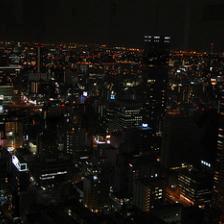}};
            \node[box] at (6, -3.2) {\includegraphics[width=2.5cm, height=2.5cm]{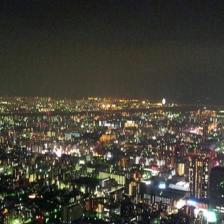}};
            \node[box] at (9, -3.2) {\includegraphics[width=2.5cm, height=2.5cm]{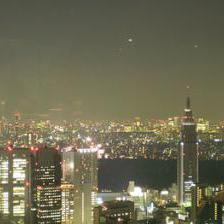}};
            \node[box] at (12, -3.2) {\includegraphics[width=2.5cm, height=2.5cm]{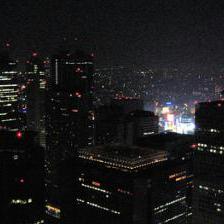}};
            \node[box] at (15, -3.2) {\includegraphics[width=2.5cm, height=2.5cm]{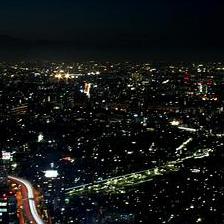}};
            \node[label style] at (-1.8, -6.4) {RTA};
            \node[box] at (0,-6.4) {\includegraphics[width=2.5cm, height=2.5cm]{suppl_eccv/reference/rta/eps4/462922891_67372e6aed_196_25159586_N00.png}};
            
            \node[box] at (3, -6.4) {\includegraphics[width=2.5cm, height=2.5cm]{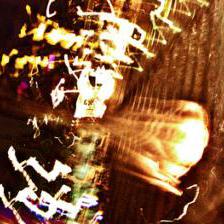}};
            \node[box] at (6, -6.4) {\includegraphics[width=2.5cm, height=2.5cm]{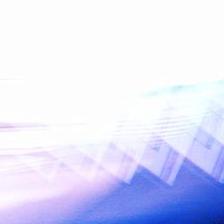}};
            \node[box] at (9, -6.4) {\includegraphics[width=2.5cm, height=2.5cm]{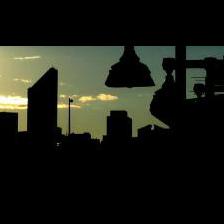}};
            \node[box] at (12, -6.4) {\includegraphics[width=2.5cm, height=2.5cm]{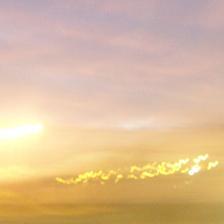}};
            \node[box] at (15, -6.4) {\includegraphics[width=2.5cm, height=2.5cm]{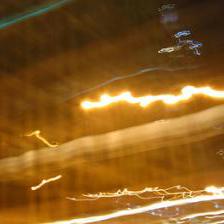}};
            \draw[thick, dotted] (1.5, 2.8) -- (1.5, -8.0);

        \end{tikzpicture}
    }
    \caption{Example of retrieved images by G3\cite{jia2024g3}. $\varepsilon = 4/255$}
    \label{fig:g3_eps4_tokyo}
\end{figure}

\begin{figure}[h]
    \centering
    \resizebox{\columnwidth}{!}{
        \begin{tikzpicture}[
            box/.style={rectangle, draw, thick, minimum size=2.5cm, inner sep=0pt},
            label style/.style={rotate=90, font=\large, anchor=south}, 
            header style/.style={font=\sffamily\Large\bfseries, anchor=south}
        ]

            \node[header style] at (0, 2.0) {Query};
            \node[header style] at (9, 2.0) {Top-5 Similar Images};

            \node[label style] at (-1.8, 0) {Ground Truth};
            \node[box] at (0,0) {\includegraphics[width=2.5cm, height=2.5cm]{suppl_eccv/reference/clean/1194828036_f64ee2efe4_1197_47311799_N00.jpg}}; 
            
            \node[box] at (3, 0) {\includegraphics[width=2.5cm, height=2.5cm]{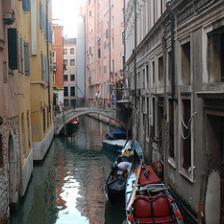}};
            \node[box] at (6, 0) {\includegraphics[width=2.5cm, height=2.5cm]{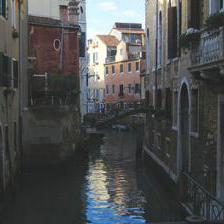}};
            \node[box] at (9, 0) {\includegraphics[width=2.5cm, height=2.5cm]{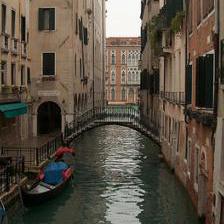}};
            \node[box] at (12, 0) {\includegraphics[width=2.5cm, height=2.5cm]{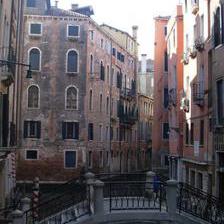}};
            \node[box] at (15, 0) {\includegraphics[width=2.5cm, height=2.5cm]{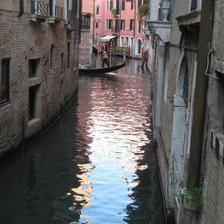}};
            \node[label style] at (-1.8, -3.2) {GeoShield\cite{geoshield}};
            \node[box] at (0,-3.2) {\includegraphics[width=2.5cm, height=2.5cm]{suppl_eccv/reference/geoshield/eps8/1194828036_f64ee2efe4_1197_47311799_N00.jpg}};
            
            \node[box] at (3, -3.2) {\includegraphics[width=2.5cm, height=2.5cm]{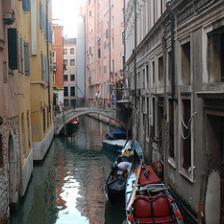}};
            \node[box] at (6, -3.2) {\includegraphics[width=2.5cm, height=2.5cm]{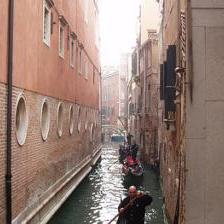}};
            \node[box] at (9, -3.2) {\includegraphics[width=2.5cm, height=2.5cm]{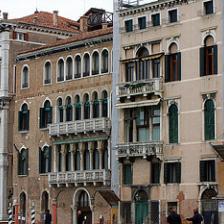}};
            \node[box] at (12, -3.2) {\includegraphics[width=2.5cm, height=2.5cm]{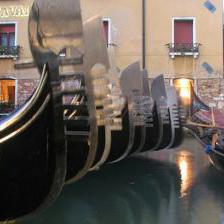}};
            \node[box] at (15, -3.2) {\includegraphics[width=2.5cm, height=2.5cm]{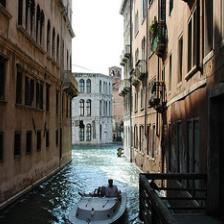}};
            \node[label style] at (-1.8, -6.4) {RTA};
            \node[box] at (0,-6.4) {\includegraphics[width=2.5cm, height=2.5cm]{suppl_eccv/reference/rta/eps8/1194828036_f64ee2efe4_1197_47311799_N00.png}};
            
            \node[box] at (3, -6.4) {\includegraphics[width=2.5cm, height=2.5cm]{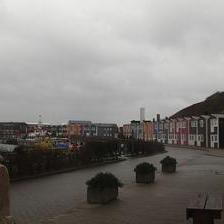}};
            \node[box] at (6, -6.4) {\includegraphics[width=2.5cm, height=2.5cm]{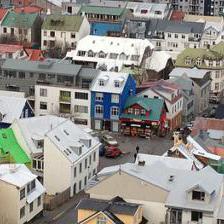}};
            \node[box] at (9, -6.4) {\includegraphics[width=2.5cm, height=2.5cm]{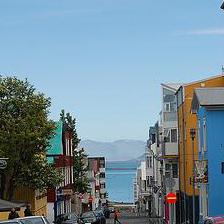}};
            \node[box] at (12, -6.4) {\includegraphics[width=2.5cm, height=2.5cm]{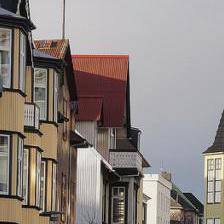}};
            \node[box] at (15, -6.4) {\includegraphics[width=2.5cm, height=2.5cm]{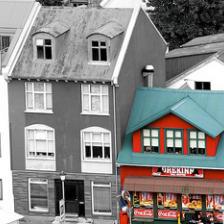}};
            \draw[thick, dotted] (1.5, 2.8) -- (1.5, -8.0);

        \end{tikzpicture}
    }
    \caption{Example of retrieved images by G3\cite{jia2024g3}. $\varepsilon = 8/255$}
    \label{fig:g3_eps8_venice}
\end{figure}

\begin{figure}[h]
    \centering
    \resizebox{\columnwidth}{!}{
        \begin{tikzpicture}[
            box/.style={rectangle, draw, thick, minimum size=2.5cm, inner sep=0pt},
            label style/.style={rotate=90, font=\large, anchor=south}, 
            header style/.style={font=\sffamily\Large\bfseries, anchor=south}
        ]

            \node[header style] at (0, 2.0) {Query};
            \node[header style] at (9, 2.0) {Top-5 Similar Images};

            \node[label style] at (-1.8, 0) {Ground Truth};
            \node[box] at (0,0) {\includegraphics[width=2.5cm, height=2.5cm]{suppl_eccv/reference/clean/1058450330_29b27f18cc_1304_52057668_N00.jpg}}; 
            
            \node[box] at (3, 0) {\includegraphics[width=2.5cm, height=2.5cm]{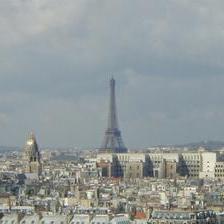}};
            \node[box] at (6, 0) {\includegraphics[width=2.5cm, height=2.5cm]{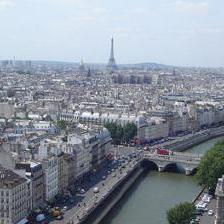}};
            \node[box] at (9, 0) {\includegraphics[width=2.5cm, height=2.5cm]{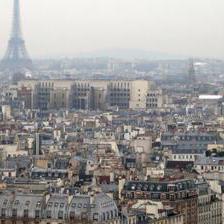}};
            \node[box] at (12, 0) {\includegraphics[width=2.5cm, height=2.5cm]{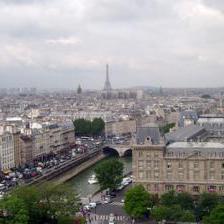}};
            \node[box] at (15, 0) {\includegraphics[width=2.5cm, height=2.5cm]{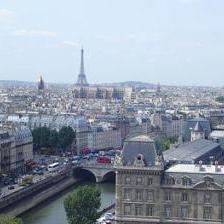}};
            \node[label style] at (-1.8, -3.2) {GeoShield\cite{geoshield}};
            \node[box] at (0,-3.2) {\includegraphics[width=2.5cm, height=2.5cm]{suppl_eccv/reference/geoshield/eps16/1058450330_29b27f18cc_1304_52057668_N00.jpg}};
            
            \node[box] at (3, -3.2) {\includegraphics[width=2.5cm, height=2.5cm]{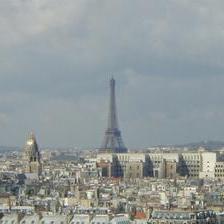}};
            \node[box] at (6, -3.2) {\includegraphics[width=2.5cm, height=2.5cm]{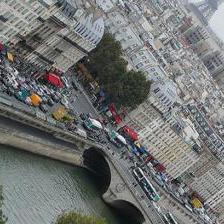}};
            \node[box] at (9, -3.2) {\includegraphics[width=2.5cm, height=2.5cm]{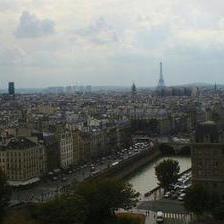}};
            \node[box] at (12, -3.2) {\includegraphics[width=2.5cm, height=2.5cm]{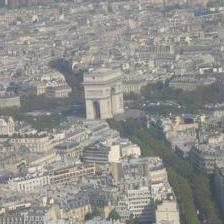}};
            \node[box] at (15, -3.2) {\includegraphics[width=2.5cm, height=2.5cm]{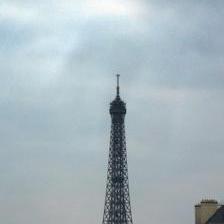}};
            \node[label style] at (-1.8, -6.4) {RTA};
            \node[box] at (0,-6.4) {\includegraphics[width=2.5cm, height=2.5cm]{suppl_eccv/reference/rta/eps16/1058450330_29b27f18cc_1304_52057668_N00.png}};
            
            \node[box] at (3, -6.4) {\includegraphics[width=2.5cm, height=2.5cm]{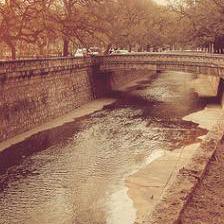}};
            \node[box] at (6, -6.4) {\includegraphics[width=2.5cm, height=2.5cm]{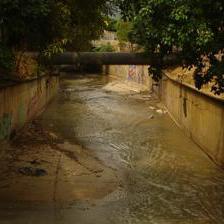}};
            \node[box] at (9, -6.4) {\includegraphics[width=2.5cm, height=2.5cm]{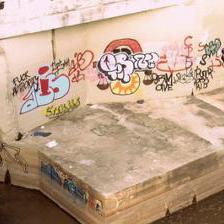}};
            \node[box] at (12, -6.4) {\includegraphics[width=2.5cm, height=2.5cm]{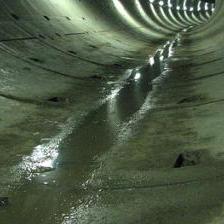}};
            \node[box] at (15, -6.4) {\includegraphics[width=2.5cm, height=2.5cm]{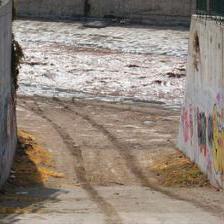}};
            \draw[thick, dotted] (1.5, 2.8) -- (1.5, -8.0);

        \end{tikzpicture}
    }
    \caption{Example of retrieved images by G3\cite{jia2024g3}. $\varepsilon = 16/255$}
    \label{fig:g3_eps16_parigi}
\end{figure}

In this section, we analyze the internal representations of the black-box target models (Img2Loc~\cite{img2loc} and G3~\cite{jia2024g3}) by examining the nearest neighbors retrieved during the retrieval stage. This qualitative assessment allows us to visualize how adversarial perturbations displace the query image within the latent manifold. 
Let $\Phi(I)$ represent the image encoding. We define the intra-modal similarity between a query image $I_q$ and a gallery image $I_g$ as:\begin{equation}\text{sim}_{I2I}(I_q, I_g) = \langle \Phi(I_q), \Phi(I_g) \rangle\end{equation}
In Figures \ref{fig:img2loc_eps2_mountain}, \ref{fig:img2loc_eps4_tokyo}, \ref{fig:img2loc_eps8_venice}, \ref{fig:img2loc_eps16_parigi}, \ref{fig:g3_eps2_mountain}, \ref{fig:g3_eps4_tokyo}, \ref{fig:g3_eps8_venice}, and \ref{fig:g3_eps16_parigi} we display the top-5 retrieved images for each attack scenario. In the baseline case of unattacked images, the visual consistency of the internal representations is evident; the top-5 nearest neighbors frequently correspond to the exact same landmark or geographic location as the query. Interestingly, this visual coherence largely persists for images attacked with GeoShield, suggesting that the perturbation is insufficient to fully decouple the image from its original semantic cluster. In contrast, for images attacked with our proposed method, 
\nn{we observe a significant representational shift.} Even at a small perturbation budget of $\varepsilon=2$, the retrieved neighbors diverge sharply from the query’s original location, successfully pushing the representation into visually distant regions of the embedding space.
\hypersetup{hidelinks}

\end{document}